\documentclass{article} 
\usepackage{iclr2027_conference,times}

\usepackage{amsmath,amssymb,amsthm,bm}

\def\eqref#1{equation~\ref{#1}}

\def\1{\bm{1}}

\DeclareMathAlphabet{\mathsfit}{\encodingdefault}{\sfdefault}{m}{sl}
\SetMathAlphabet{\mathsfit}{bold}{\encodingdefault}{\sfdefault}{bx}{n}

\theoremstyle{plain}
\newtheorem{theorem}{Theorem}[section]

\newtheorem{lemma}[theorem]{Lemma}

\theoremstyle{definition}

\newtheorem{assumption}[theorem]{Assumption}
\theoremstyle{remark}

\usepackage{hyperref}
\usepackage{url}

\usepackage[T1]{fontenc}
\usepackage{algorithmic}
\usepackage{algorithm}
\usepackage{tcolorbox}
\usepackage{graphicx}
\usepackage{bbding}
\usepackage{subcaption}
\usepackage{booktabs}
\usepackage{multirow}
\usepackage{titletoc}
\renewcommand{\cite}{\citep}
\usepackage{color, colortbl}
\definecolor{blue1}{rgb}{0.6, 0.6, 0.6}
\definecolor{blue2}{rgb}{0.75, 0.8, 0.99}
\definecolor{blue3}{rgb}{0.84,0.89,0.99}
\definecolor{blue4}{rgb}{0.93,0.95,0.99}

\definecolor{DarkPink}{rgb}{0.5,0.0,0.18}
\definecolor{DarkGreen}{rgb}{0.1,0.5,0.1}
\definecolor{DarkRed}{rgb}{0.5,0.1,0.1}
\definecolor{DarkBlue}{rgb}{0.1,0.1,0.7}
\definecolor{DarkYellow}{rgb}{.79,.79,0}
\hypersetup{
    unicode=false,          
    pdftoolbar=true,        
    pdfmenubar=true,        
    pdffitwindow=false,      
    pdfnewwindow=true,      
    colorlinks=true,       
    linkcolor=DarkBlue,          
    citecolor=DarkRed,        
    filecolor=DarkRed,      
    urlcolor=DarkBlue,          
    pdftitle={},
    pdfauthor={},
}

\title{GWT: Scalable Optimizer State Compression for Large Language Model Training}

\iclrpreprintcopy

\author{
\textbf{
Ziqing Wen\textsuperscript{1},
Ping Luo\textsuperscript{1}, 
Jiahuan Wang\textsuperscript{1}, 
Junlin Zeng\textsuperscript{1}, 
Kun Yuan\textsuperscript{2}, 
}
\\
\textbf{
\
Dongsheng Li\textsuperscript{1},
and Tao Sun\textsuperscript{1}\thanks{Correspondence to: Tao Sun \{suntao.saltfish@outlook.com\}}
}
\\[4pt]
\ \textsuperscript{1} College of Computer Science and Technology, National University of Defense Technology
\\
\ \textsuperscript{2} Center for Machine Learning Research, Peking University
}

\begin{document}

\maketitle

\begin{abstract}
Training large language models (LLMs) requires substantial memory, a significant fraction of which is consumed by the moment states maintained by adaptive optimizers such as Adam. Existing memory-efficient approaches commonly compress parameters or gradients through low-rank projections, which may discard information outside the selected subspace or incur additional costs from subspace construction and singular value decomposition (SVD). We introduce \textbf{Gradient Wavelet Transform (GWT)}, an SVD-free framework that applies a multilevel Haar transform to matrix gradients. GWT maintains first- and second-moment states only for the compact approximation coefficients, while retaining all detail coefficients transiently in each update. We show that the orthogonal transform preserves the complete current gradient and establish gradient alignment and a stationary-point guarantee for smooth objectives under bounded preconditioning and a directly verifiable coarse-momentum coherence condition. Experiments on language-model pre-training and downstream fine-tuning show that GWT reduces estimated model-and-optimizer-state memory by up to 52\% while maintaining competitive or improved model quality. On LLaMA-3B pre-training, GWT achieves $1.9\times$ the throughput of 8-bit Adam and is compatible with multiple optimizer families, including Adam, Adam-mini, and Muon.
\end{abstract}

\section{Introduction}

\begin{figure*}[!t]
\centering
\includegraphics[width=\linewidth]{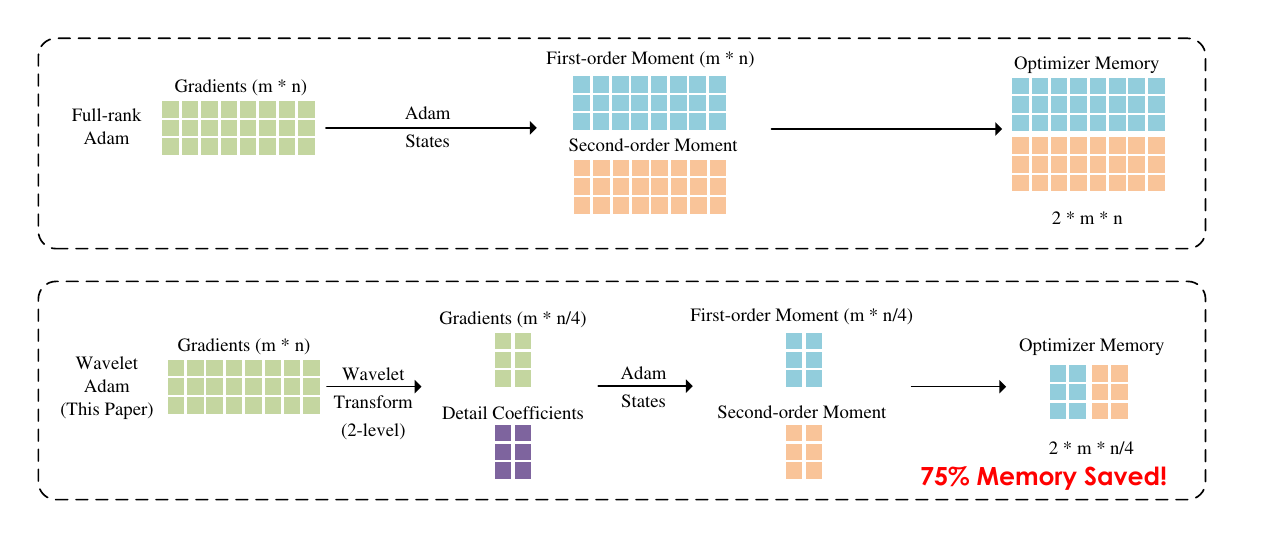}
    \caption{\textbf{Persistent optimizer-state memory under GWT.} For a matrix updated with GWT, a 2-level Haar transform stores Adam states for only 25\% of the coefficients.}\label{fig:visualization_memory_usage}
\end{figure*}

Training large language models (LLMs) requires substantial GPU memory, a considerable fraction of which is consumed by optimizer states. Adam-style optimizers~\cite{Kingma2014AdamAM}, despite their strong convergence and robustness, maintain first- and second-moment estimates for each trainable parameter. Consequently, their optimizer states can require twice the memory of the model parameters. For example, under the BF16 configuration illustrated in Figure~\ref{fig:visualization_memory_usage}, pre-training a 7B-parameter model requires approximately 28\,GB solely for Adam states~\cite{zhao2024galore}. This overhead has become a major bottleneck in scaling LLM training.

Existing memory-efficient training methods largely rely on low-rank representations. LoRA~\cite{hu2021lora} reduces trainable parameters by constraining weight updates to low-rank factors, but this constraint can limit its applicability to pre-training and other demanding optimization settings~\cite{xia2024chainlora,zhang2023lorafa}. GaLore~\cite{zhao2024galore} instead projects full-Adam gradients into low-dimensional subspaces and maintains optimizer states in the compressed space. Subsequent methods compensate for projection residuals~\cite{Chen2024FiraCW}, improve robustness to noisy gradients~\cite{he2024subspacegolore}, or reduce the cost of subspace construction and updates~\cite{zhu2024apollosgdlikememoryadamwlevel_apollo,robert2025ldadam}. Nevertheless, global low-rank representations organize gradients according to global spectral structure and may discard components outside the selected subspace. This motivates investigating whether a localized, multiresolution representation can provide a different structure for optimizer-state compression.

In particular, wavelet transforms separate a signal into coarse approximation coefficients and localized detail coefficients while preserving the complete signal through exact reconstruction. This raises the following question:
\begin{tcolorbox}[
    colback=pink!30,
    colframe=pink!60!black,
    boxrule=0.8pt,
    arc=6pt,
    left=6pt, right=6pt, top=6pt, bottom=6pt
]
\begin{center}
  \textit{Can the multiresolution structure of wavelets be used to compress optimizer states while preserving the complete instantaneous gradient?}
\end{center}
\end{tcolorbox}

Wavelet transforms are well suited to this objective. They represent signals through localized components at multiple resolutions and admit linear-complexity transforms in the number of matrix elements~\cite{Wavelets_and_signal}. When applied to matrix gradients, the coarse approximation coefficients capture block-wise aggregate structure, whereas the detail coefficients retain variations at finer scales. This decomposition makes it possible to maintain persistent states only for the compact approximation coefficients, while using the current detail coefficients directly during each update.

Motivated by this observation, we introduce \textbf{Gradient Wavelet Transform (GWT)}, a plug-and-play framework that applies a multilevel Haar transform to matrix gradients. GWT stores optimizer states only for the coarsest approximation coefficients and reconstructs each update using both the stateful approximation coefficients and the current detail coefficients. Unlike methods that retain only a projected gradient, GWT preserves all instantaneous Haar coefficients and compresses only the temporally accumulated optimizer states. It requires neither repeated SVD nor persistent projection matrices and can be integrated with multiple optimizer families. Empirically, we find that the compact approximation band captures substantially more gradient energy than its dimensional fraction, with this concentration weakened by random column permutation, indicating exploitable local gradient structure. Moreover, the reconstructed GWT direction remains positively aligned with the instantaneous gradient across all sampled cases, while the retained detail coefficients provide a substantial positive contribution to the update. In our experiments, GWT reduces estimated model-and-optimizer-state memory by up to 52\% and achieves up to $1.9\times$ the throughput of 8-bit Adam on LLaMA-3B pre-training while maintaining competitive model quality.

Our main contributions are summarized as follows:
\begin{itemize}

\item We propose \textbf{Gradient Wavelet Transform (GWT)}, a wavelet-domain
optimizer-state compression framework that maintains adaptive states only
for the compact Haar approximation coefficients while retaining all
instantaneous detail coefficients during reconstruction. GWT requires
neither repeated SVD nor persistent projection matrices and can be readily
integrated with different stateful optimizers.

\item We provide theoretical and empirical analyses to characterize the
mechanism underlying GWT. We show that the Haar transform exactly preserves
the instantaneous gradient, relate local column variation to the energy of
the detail coefficients, and characterize the temporal error introduced by
maintaining momentum only in the approximation band. Under bounded
preconditioning and coarse-momentum coherence, the reconstructed GWT
direction remains gradient-aligned and admits a convergence guarantee.
Our empirical diagnostics further show that the approximation band captures
disproportionate gradient energy, exploits non-random local gradient
structure, and works together with the retained detail coefficients to
preserve effective update directions.

\item We extensively evaluate GWT across language-model pre-training and
downstream fine-tuning. GWT substantially reduces optimizer-state memory
and improves training throughput while maintaining competitive model
quality. It also remains effective across different decomposition levels,
model scales, and optimizer families.

\end{itemize}

\section{Related Work}

\textbf{Memory-Efficient Training and Optimization.}
Parameter-efficient methods reduce training memory by restricting the number or structure of trainable parameters. LoRA~\cite{hu2021lora}, for example, freezes pretrained weights and represents updates with trainable low-rank factors. Numerous extensions improve its optimization dynamics, expressiveness, or applicability to continued training~\cite{hao2024flora,Lialin2023ReLoRAHT,xia2024chainlora,zhang2023lorafa,liu2024dora,luo2024badam}. For full-parameter training, GaLore~\cite{zhao2024galore} projects gradients into low-dimensional subspaces and maintains optimizer states in the compressed space. Subsequent methods compensate for discarded information, reduce subspace-update overhead, or improve compressed optimization~\cite{Chen2024FiraCW,zhu2024apollosgdlikememoryadamwlevel_apollo,robert2025ldadam,rajabi2025subtrack++}. Another line of work reduces optimizer states through matrix orthogonalization or gradient normalization, including Muon~\cite{liu2025muon}, SWAN~\cite{Ma2024SWANSW}, SinkGD~\cite{scetbon2025sinkgd}, and SCALE~\cite{glentis2025a_minimalist}. GWT differs from these methods by using a fixed multilevel transform rather than a data-dependent low-rank factorization or repeated SVD.

\textbf{Transform-Domain Compression.}
Wavelet transforms provide localized multiresolution representations and have long been used for efficient signal and image compression~\cite{Porwik2004ThehaarT,Wavelets_and_signal}. GWT transfers this principle to optimizer-state compression: the transform itself is lossless, but persistent adaptive states are allocated only to the approximation coefficients, while detail coefficients are retained transiently during reconstruction.

\textbf{System-Level Memory Efficiency.}
System-level techniques target memory components beyond optimizer states. Activation checkpointing~\cite{chen2016training} reduces activation memory by recomputing intermediate values during the backward pass, whereas quantization~\cite{dettmers20218bit,dettmers2024qloraquantized} reduces storage or computation through lower-precision representations. These techniques address complementary memory sources and can therefore be combined with GWT.

\section{Methods}\label{sec:method}

We first define the multilevel Haar representation used by GWT and then describe how adaptive optimizer states are maintained in the transformed domain. Throughout this section, $t$ denotes the optimization iteration and $j$ denotes the wavelet level. Implementation details, an eight-dimensional example, compact pseudocode, and the Norm-growth Limiter are provided in Appendix~\ref{app:implementation_details}.

\subsection{Multilevel Haar Gradient Representation}
Let $G_t\in\mathbb{R}^{m\times n}$ be the matrix gradient used at iteration $t$, where $2^l$ divides $n$, and set $A_t^{(0)}=G_t$. At level $j\in\{1,\ldots,l\}$, a one-level orthonormal Haar transform is applied along the column dimension:
\begin{equation}\nonumber
    [A_t^{(j)},D_t^{(j)}]
    =A_t^{(j-1)}H_{n/2^{j-1}},
\end{equation}
where $A_t^{(j)},D_t^{(j)}\in\mathbb{R}^{m\times n/2^j}$ are the approximation and detail coefficients, respectively. For an even dimension $s$, the Haar matrix $H_s\in\mathbb{R}^{s\times s}$ is orthogonal and satisfies
\begin{equation}\label{eq:A_D_define}
    (H_s)_{2i-1,i}=(H_s)_{2i,i}=(H_s)_{2i-1,s/2+i}=\frac{1}{\sqrt{2}},
    \qquad
    (H_s)_{2i,s/2+i}=-\frac{1}{\sqrt{2}},
\end{equation}
for $i=1,\ldots,s/2$, with all remaining entries equal to zero. The complete level-$l$ transform is
\begin{equation}\label{eq:multilevel_haar}
    \mathcal{H}_l(G_t)
    =\big(A_t^{(l)},D_t^{(l)},D_t^{(l-1)},\ldots,D_t^{(1)}\big),
\end{equation}
and orthogonality gives the exact reconstruction
\begin{equation}\label{eq:multilevel_haar_inverse}
    G_t=\mathcal{H}_l^{-1}
    \big(A_t^{(l)},D_t^{(l)},\ldots,D_t^{(1)}\big).
\end{equation}
It also gives the Parseval identity
\begin{equation}\label{eq:haar_parseval}
    \|G_t\|_F^2
    =\|A_t^{(l)}\|_F^2
    +\sum_{j=1}^{l}\|D_t^{(j)}\|_F^2.
\end{equation}
Thus, the transform preserves all current gradient information, while the coarsest approximation contains only $mn/2^l$ entries. GWT uses this asymmetry to compress the persistent optimizer states rather than the current gradient.

\subsection{Adam with GWT}
For a matrix $W_t\in\mathbb{R}^{m\times n}$, let $\beta_1,\beta_2\in[0,1)$ and $\epsilon>0$. Full Adam updates its first- and second-moment states using
\begin{equation}\label{eq:adam_STATE_update}
\begin{aligned}
M_{t+1} &= \beta_1M_t+(1-\beta_1)G_t,
V_{t+1} = \beta_2V_t+(1-\beta_2)G_t^2,
\end{aligned}
\end{equation}
where the square is element-wise. These states contain $2mn$ persistent entries.

GWT instead maintains moments only for the coarsest approximation coefficients:
\begin{equation}\label{eq:gwt_state_update}
\begin{aligned}
M_{t+1}^{A}
&=\beta_1M_t^{A}+(1-\beta_1)A_t^{(l)},\ 
V_{t+1}^{A}
=\beta_2V_t^{A}+(1-\beta_2)(A_t^{(l)})^2,\\
\widehat M_{t+1}^{A}
&=\frac{M_{t+1}^{A}}{1-\beta_1^{t+1}},
\qquad
\widehat V_{t+1}^{A}
=\frac{V_{t+1}^{A}}{1-\beta_2^{t+1}}.
\end{aligned}
\end{equation}
Let $\mathcal{B}_{j,l}$ be the dyadic shape-alignment operator that expands a coarsest-level tensor to the shape of $D_t^{(j)}$ over the corresponding Haar supports. We define the positive element-wise preconditioners
\begin{equation}\label{eq:gwt_preconditioners}
\begin{aligned}
S_{t+1}^{A}
&=\frac{1}{\sqrt{\widehat V_{t+1}^{A}}+\epsilon},\ 
S_{t+1}^{D,j}
=\frac{1}{\sqrt{\mathcal{B}_{j,l}(\widehat V_{t+1}^{A})}+\epsilon},
\qquad j=1,\ldots,l.
\end{aligned}
\end{equation}
The transformed update coefficients are
\begin{equation}\label{eq:gwt_coefficient_update}
\begin{aligned}
\widetilde A_t^{(l)}
=S_{t+1}^{A}\odot\widehat M_{t+1}^{A},\
\widetilde D_t^{(j)}
=S_{t+1}^{D,j}\odot D_t^{(j)},
\qquad j=1,\ldots,l,
\end{aligned}
\end{equation}
where $\odot$ denotes element-wise multiplication. The GWT direction and parameter update are then
\begin{equation}\label{eq:adam_haar}
\begin{aligned}
\widetilde G_t
&=\alpha\,\mathcal{H}_l^{-1}
\big(\widetilde A_t^{(l)},\widetilde D_t^{(l)},\ldots,\widetilde D_t^{(1)}\big), \ 
W_{t+1}=W_t-\eta\widetilde G_t,
\end{aligned}
\end{equation}
where $\alpha>0$ is the module-wise scale factor and $\eta>0$ is the learning rate. The states $M_t^A$ and $V_t^A$ persist across iterations, whereas the detail coefficients are generated and consumed within the current iteration. All square roots, divisions, and powers in ~\eqref{eq:gwt_state_update}--\eqref{eq:gwt_coefficient_update} are element-wise.

For training stability, we additionally adopt the Norm-growth Limiter of Fira~\cite{Chen2024FiraCW} with $\gamma=1.01$. Its definition and ablation are deferred to Appendix~\ref{app:norm_growth_limiter}, since it is an auxiliary stabilization technique rather than a component specific to the wavelet transform.

\subsection{Memory and Computational Complexity}
Full Adam stores two persistent states of size $mn$. GWT stores two states of size $mn/2^l$, giving
\begin{equation}\nonumber
    \mathrm{StateMem}_{\mathrm{GWT}-l}=\frac{2mn}{2^l},
    \qquad
    \mathrm{Reduction}=1-2^{-l}.
\end{equation}
This is the reduction for a GWT-treated matrix; whole-model savings are smaller when embeddings, normalization parameters, or other modules continue to use full Adam. The detail coefficients do not contribute persistent optimizer-state memory, although the transform requires temporary coefficient buffers during the current iteration.

A multilevel Haar transform processes progressively shorter approximation bands. Its arithmetic cost is
\begin{equation}\nonumber
    \mathcal{O}\!\left(mn+\frac{mn}{2}+\cdots+\frac{mn}{2^{l-1}}\right)
    =\mathcal{O}(mn),
\end{equation}
and the inverse transform has the same order. Updating the compressed moment states costs $\mathcal{O}(mn/2^l)$, while preconditioning the detail bands and reconstructing the update cost $\mathcal{O}(mn)$. Therefore, the overall GWT update remains linear in the number of matrix entries and requires neither SVD nor a persistent projection matrix.

\begin{table*}[!t]
    \centering
    \caption{\textbf{Persistent optimizer-state memory and amortized
    per-step complexity for one matrix $W\in\mathbb{R}^{m\times n}$,
    assuming $m\leq n$.}
    $r$ is the projection rank.
    }
    \label{tab:memory_theory_comp}
    \resizebox{0.9\linewidth}{!}{
    \begin{tabular}{lccc}
        \toprule
        Method
        & Optimizer states
        & Projection storage
        & Update complexity \\
        \midrule
        Full Adam
        & $2mn$
        & $0$
        & $\mathcal{O}(mn)$ \\

        Muon~\cite{liu2025muon}
        & $mn$
        & $0$
        & $\mathcal{O}(m^{2}n)$ \\

        GaLore~\cite{zhao2024galore}
        & $2nr$
        & $mr$
        & $\mathcal{O}\!\left(
            m^{2}n
          \right)$ \\
        APOLLO~\cite{zhu2024apollosgdlikememoryadamwlevel_apollo}
        & $2nr$
        & Seed only $(\mathcal{O}(1))$
        & $\mathcal{O}(mnr)$ \\
        GWT-$l$
        & ${2mn}/{2^l}$
        & $0$
        & $\mathcal{O}(mn)$ \\
        \bottomrule
    \end{tabular}
    }
\end{table*}

\subsection{Theoretical Analysis}\label{sec:theory}
The analysis follows the structure of the algorithm. We first show that the Haar transform preserves the complete current gradient and quantify the energy of its detail coefficients. We then relate the approximation-state momentum error to temporal changes in the coarse coefficients. These two results identify the spatial and temporal structures used by GWT and lead to a descent and stationary-point convergence guarantee for the reconstructed direction.

\paragraph{Information preservation and detail energy.}
~\eqref{eq:multilevel_haar_inverse} shows that the Haar transform itself introduces no approximation error. GWT retains every current detail coefficient and reduces memory only by compressing the temporal optimizer states. Let $P_l(G)$ denote the reconstruction obtained by retaining $A^{(l)}$ and setting all detail bands to zero, and define
\begin{equation}\nonumber
    (\Delta G)_{:,k}=G_{:,k+1}-G_{:,k},
    \qquad k=1,\ldots,n-1.
\end{equation}

\begin{lemma}[Detail-Energy Bound]\label{lem:detail_energy}
Let $b=2^l$ divide $n$ and define $\kappa_b=[2\sin(\pi/(2b))]^{-1}$. Then
\begin{equation}\label{eq:detail_energy_bound}
    \sum_{j=1}^{l}\|D^{(j)}\|_F^2
    =\|G-P_l(G)\|_F^2
    \leq \kappa_b^2\|\Delta G\|_F^2.
\end{equation}
\end{lemma}
The bound is not required for GWT to operate. It states that when adjacent columns vary slowly, a larger share of the gradient energy lies in the approximation coefficients. The remaining detail information is nevertheless retained in the reconstructed direction in ~\eqref{eq:adam_haar}.

\paragraph{Temporal coherence of the approximation state.}
The approximation coefficients are the only coefficients for which GWT stores a first-moment state. Define the coarse-momentum error
\begin{equation}\nonumber
    \Xi_t^A=\widehat M_{t+1}^A-A_t^{(l)}.
\end{equation}

\begin{lemma}[Temporal Form of the Coarse-Momentum Error]\label{lem:momentum_temporal}
Assume $M_0^A=0$ and $0\leq\beta_1<1$. For $q=0,\ldots,t$ and $u=0,\ldots,t-1$, define
\begin{equation}\nonumber
    \omega_{t,q}
    =\frac{(1-\beta_1)\beta_1^{t-q}}{1-\beta_1^{t+1}},
    \qquad
    c_{t,u}
    =\frac{\beta_1^{t-u}(1-\beta_1^{u+1})}{1-\beta_1^{t+1}}.
\end{equation}
Then
\begin{equation}\label{eq:temporal_momentum_error}
\begin{aligned}
    \widehat M_{t+1}^A
    =\sum_{q=0}^{t}\omega_{t,q}A_q^{(l)},\ 
    \Xi_t^A
    =-\sum_{u=0}^{t-1}c_{t,u}
    \big(A_{u+1}^{(l)}-A_u^{(l)}\big).
\end{aligned}
\end{equation}
Consequently,
\begin{equation}\label{eq:temporal_momentum_bound}
    \left\|(S_{t+1}^{A})^{1/2}\odot\Xi_t^A\right\|_F
    \leq
    \sum_{u=0}^{t-1}c_{t,u}
    \left\|(S_{t+1}^{A})^{1/2}\odot
    \big(A_{u+1}^{(l)}-A_u^{(l)}\big)\right\|_F.
\end{equation}
For $t=0$, both sides of ~\eqref{eq:temporal_momentum_bound} are zero.
\end{lemma}
Thus, the approximation-state error is small when the coarse coefficients vary gradually across nearby iterations. This relation also provides a direct empirical diagnostic for the assumption used below.

\begin{assumption}[Bounded Preconditioning and Coarse-Momentum Coherence]\label{assump:gwt_alignment}
There exist constants $0<s_{\min}\leq s_{\max}<\infty$ and $0\leq\rho<1$ such that, for every iteration $t\in\{0,\ldots,T-1\}$ and level $j\in\{1,\ldots,l\}$, each entry of $S_{t+1}^{A}$ and $S_{t+1}^{D,j}$ lies in $[s_{\min},s_{\max}]$, and
\begin{equation}\label{eq:coarse_coherence}
    \left\|(S_{t+1}^{A})^{1/2}\odot\Xi_t^A\right\|_F
    \leq
    \rho\left\|(S_{t+1}^{A})^{1/2}\odot A_t^{(l)}\right\|_F.
\end{equation}
\end{assumption}
The preconditioner bounds control the effective coordinate-wise step sizes. Since $\epsilon>0$, the entries are automatically bounded above by $1/\epsilon$; a positive lower bound follows whenever the second-moment estimates remain uniformly bounded. The coherence condition controls only the stateful approximation band and imposes no smallness condition on the detail coefficients.

\begin{theorem}[Descent and Convergence of GWT]\label{thm:gwt_descent}
Let $F$ be lower bounded by $F_*$ and have an $L$-Lipschitz gradient with respect to the Frobenius norm. For $t=0,\ldots,T-1$, let $G_t=\nabla F(W_t)$ and construct $\widetilde G_t$ using ~\eqref{eq:gwt_state_update}--\eqref{eq:adam_haar}. Suppose Assumption~\ref{assump:gwt_alignment} holds with the same constants over these iterations, and define
\begin{equation}\nonumber
    \chi=\frac{s_{\max}}{s_{\min}},
    \qquad
    \mu=\alpha(1-\rho)s_{\min},
    \qquad
    C=\alpha s_{\max}(1+\rho\sqrt{\chi}).
\end{equation}
Then
\begin{equation}\label{eq:gwt_alignment_bound}
    \langle G_t,\widetilde G_t\rangle_F
    \geq \mu\|G_t\|_F^2,
    \qquad
    \|\widetilde G_t\|_F\leq C\|G_t\|_F.
\end{equation}
For a fixed learning rate satisfying $0<\eta<2\mu/(LC^2)$, the iterates in ~\eqref{eq:adam_haar} further satisfy
\begin{equation}\label{eq:gwt_convergence}
    \frac{1}{T}\sum_{t=0}^{T-1}\|\nabla F(W_t)\|_F^2
    \leq
    \frac{F(W_0)-F_*}
    {T\eta\left(\mu-\frac{L\eta C^2}{2}\right)}.
\end{equation}
\end{theorem}
The theorem shows that state compression preserves a descent direction when the coarse momentum remains coherent. The current detail coefficients always contribute a positively preconditioned component, while the approximation band supplies the temporally accumulated state.

\paragraph{Scope of the analysis.}
Theorem~\ref{thm:gwt_descent} isolates the effect of GWT state compression under full-gradient updates. Extending the result to minibatch training would require controlling the conditional bias introduced by the interaction between stochastic gradients and the adaptive wavelet-domain preconditioner; unbiased minibatch gradients alone do not imply an unbiased GWT direction. We therefore do not impose an additional stochastic-direction assumption. Instead, the experiments can directly examine the coherence ratio in ~\eqref{eq:coarse_coherence} during minibatch LLM training. The theorem analyzes the GWT direction before the optional Norm-growth Limiter. The limiter rescales the direction by a positive scalar and hence preserves its orientation, although its time-varying magnitude is not included in the stated rate. Complete proofs of Lemmas~\ref{lem:detail_energy} and~\ref{lem:momentum_temporal} and Theorem~\ref{thm:gwt_descent} are provided in Appendix~\ref{app:theory}.

\section{Experiments}\label{sec:results}

In this section, we demonstrate that the Gradient Wavelet Transform (GWT), as a memory-efficient optimization technique, can achieve performance comparable to or even surpassing that of full-Adam optimizers, while significantly reducing memory usage and increasing training throughput. Specifically, we evaluate GWT in two contexts: pre-training LLaMA \cite{Touvron2023LLaMAOA} models on the Colossal Clean Crawled Corpus (C4) \cite{2019c4} English benchmark, fine-tuning various pre-trained models on the Multi-task Language Understanding (MMLU) \cite{hendrycks2020measuringmmlu} and RoBERTa-base on the  General Language Understanding Evaluation (GLUE) \cite{wang2018glue} benchmarks. We adopt the BF-16 format to reduce memory usage across all experiments and provide a detailed description of these datasets in the Appendix.

For GWT, we use the discrete Haar wavelet as the default filter. Additionally, we present an ablation study to examine the impact of key hyperparameters, including the scale factor ($\alpha$), initial learning rate ($lr$), the GWT level ($l$), and generalization to models beyond LLaMA. A comprehensive description of the experimental setup can be found in the Appendix.

\begin{table*}[!th]
    \centering
    \caption{{Final validation perplexity (PPL, lower is better, 3 independent runs for 60M to 350M models) and estimated memory usage (BF16, model, and optimizer states) for pre-training LLaMA models on the C4 dataset.}}
    \label{tab:validation_60m-1b}
    \resizebox{\linewidth}{!}{\begin{tabular}{l|cc|cc|cc|cc}
    \toprule
        \multirow{2}{*}{Methods} & \multicolumn{2}{c|}{\textbf{60M}} & \multicolumn{2}{c|}{\textbf{130M}} & \multicolumn{2}{c|}{\textbf{350M}} & \multicolumn{2}{c}{\textbf{1.3B}} \\
        & PPL & Mem. & PPL & Mem. & PPL & Mem. & PPL & Mem. \\
        \midrule
        Full-Adam & 33.37$_{\pm 0.04}$ & 0.34G & 25.08$_{\pm 0.03}$ & 0.80G & 18.75$_{\pm 0.00}$ & 2.20G & 16.10$_{\pm 0.00}$ & 8.03G \\
        Muon & 28.93$_{\pm 0.03}$ & 0.30G & 23.05$_{\pm 0.01}$ & 0.63G & 16.96$_{\pm 0.00}$ & 1.60G & 14.28$_{\pm 0.00}$ & 5.61G\\
        Adam-Mini & 29.63$_{\pm 0.05}$ & 0.22G & 23.73$_{\pm 0.01}$ & 0.53G & 17.83$_{\pm 0.01}$ & 1.46G & 15.10$_{\pm 0.00}$ & 5.35G \\
        LDAdamW & 29.27$_{\pm 0.07}$ & 0.28G & 22.86$_{\pm 0.02}$ & 0.57G & 17.53$_{\pm 0.01}$& 1.38G & 14.88$_{\pm 0.00}$ & 4.76G \\
        \midrule
        GaLore-1/4 & 39.94$_{\pm 0.07}$ & 0.28G & 26.47$_{\pm 0.04}$ & 0.58G & 19.36$_{\pm 0.02}$ & 1.38G & 15.66$_{\pm 0.00}$ & 4.76G \\
        APOLLO-1/4 & 31.53$_{\pm 0.12}$ & 0.28G & 23.35$_{\pm 0.16}$ & 0.58G & 16.73$_{\pm 0.05}$ & 1.38G & 14.10$_{\pm 0.00}$ & 4.76G \\
        \rowcolor{blue4}\textbf{GWT-2} &  \textbf{29.35}$_{\pm 0.04}$ & 0.27G & \textbf{22.47}$_{\pm 0.04}$ & 0.55G & \textbf{16.29}$_{\pm 0.01}$ & 1.24G & \textbf{13.50}$_{\pm 0.00}$ & 4.41G\\
        \midrule
        GaLore-1/8 & 48.48$_{\pm 1.44}$ & 0.26G & 30.02$_{\pm 0.79}$ & 0.53G& 21.59$_{\pm 0.06}$ & 1.23G & 17.52$_{\pm 0.00}$ & 4.15G\\
        APOLLO-1/8 & 32.50$_{\pm 0.09}$ & 0.26G & 23.74$_{\pm 0.17}$ & 0.53G & 16.98$_{\pm 0.00}$ & 1.23G & 14.32$_{\pm 0.00}$ & 4.15G\\
        \rowcolor{blue4}\textbf{GWT-3} & \textbf{29.81}$_{\pm 0.09}$ & 0.26G & \textbf{22.63}$_{\pm 0.04}$ & 0.51G& \textbf{16.35}$_{\pm 0.02}$ & 1.09G& \textbf{13.48}$_{\pm 0.00}$ & 3.80G\\
        \midrule
        Training Tokens & \multicolumn{2}{|c}{1.3B} & \multicolumn{2}{|c}{2.6B} & \multicolumn{2}{|c}{7.8B} & \multicolumn{2}{|c}{13.1B} \\
         \bottomrule
    \end{tabular}}
\end{table*}

\subsection{Memory-efficient Pre-Training}

We integrate GWT with Adam~\cite{Kingma2014AdamAM} and evaluate it on LLaMA~\cite{Touvron2023LLaMAOA} pre-training over C4. We compare against our reproduced results for full-Adam, Adam-mini~\cite{zhang2024adammini}, LDAdamW~\cite{robert2025ldadam}, GaLore~\cite{zhao2024galore}, and APOLLO~\cite{zhu2024apollosgdlikememoryadamwlevel_apollo}. Unless otherwise specified, all experiments use cosine learning-rate decay, a sequence length of 256, and a batch size of 512, corresponding to 131K tokens per batch. GaLore and APOLLO are evaluated with ranks equal to one-fourth and one-eighth of the matrix dimension, denoted by GaLore/APOLLO-1/4 and GaLore/APOLLO-1/8. These compression ratios correspond to the state sizes of GWT-2 and GWT-3, respectively. Following prior work~\cite{zhao2024galore,zhu2024apollosgdlikememoryadamwlevel_apollo,Chen2024FiraCW}, compressed optimization is applied to the MLP and attention matrices, while the remaining parameters are updated with Adam.

\begin{figure*}[!th]
    \centering
    \subfloat[\small LLaMA 60M]{\includegraphics[width=0.33\linewidth,height=0.22\textwidth]{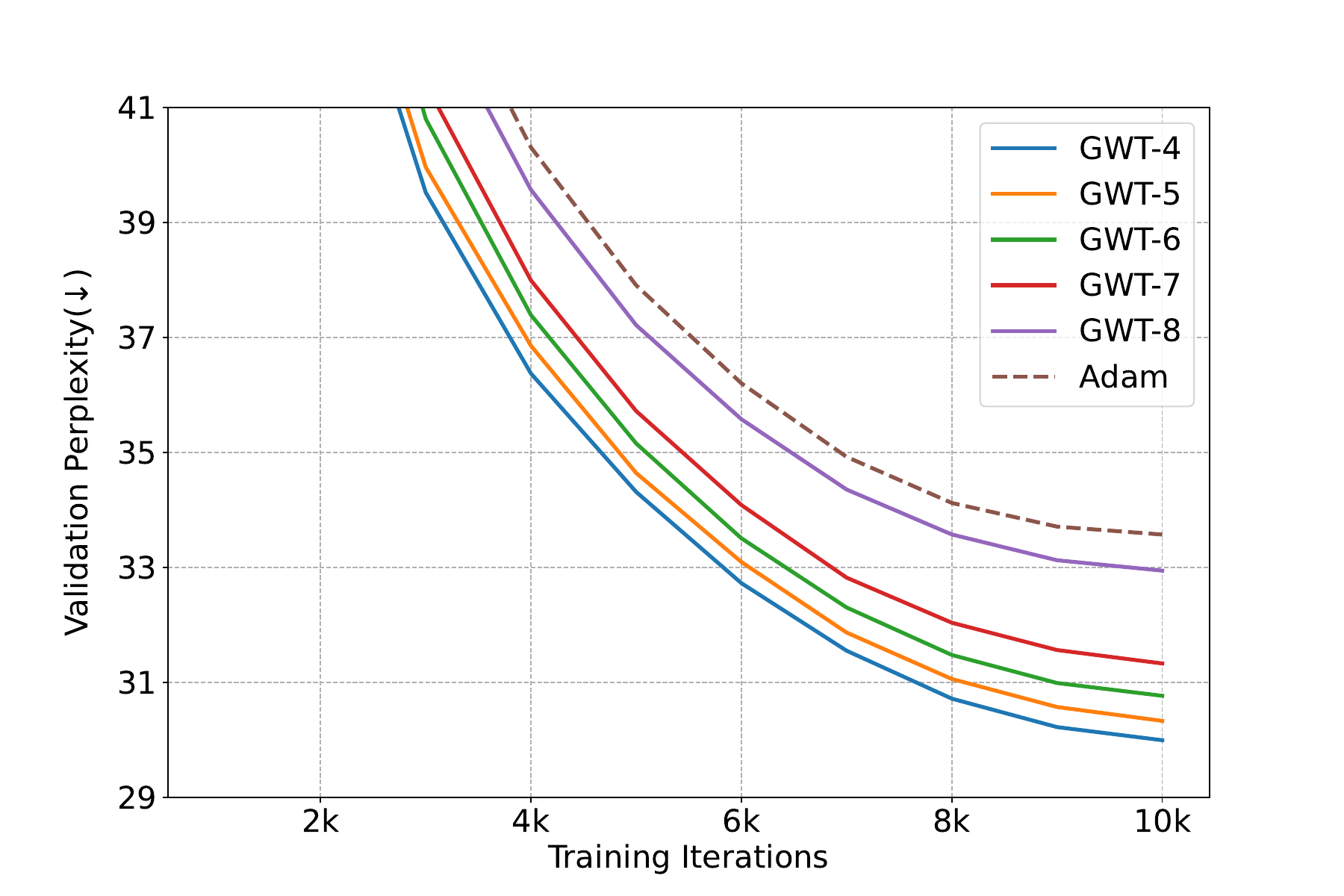}}
    \subfloat[\small LLaMA 130M]{\includegraphics[width=0.33\linewidth,height=0.22\textwidth]{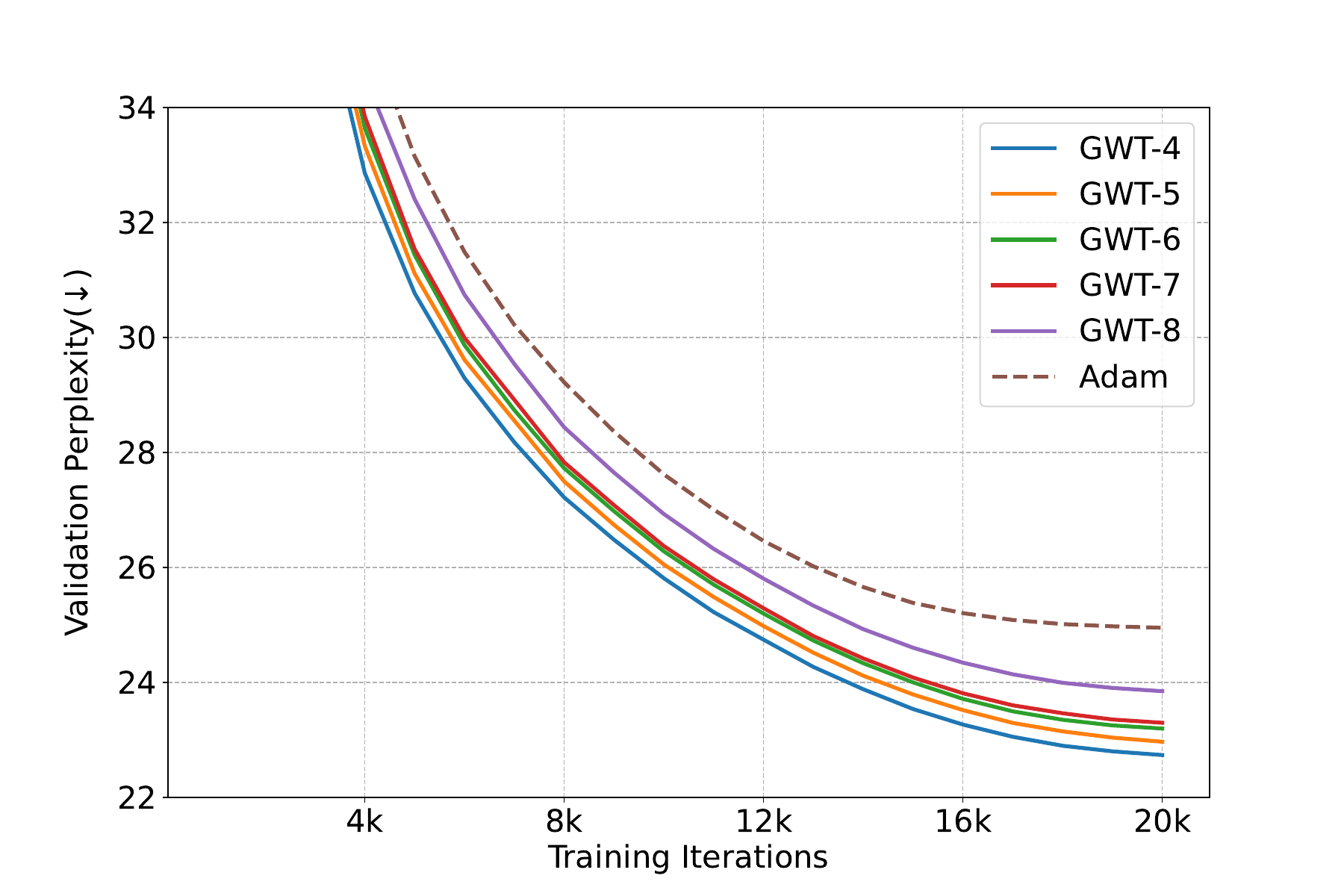}}
    \subfloat[\small LLaMA 350M]{\label{fig:learning_curve_2_adam_350m}\includegraphics[width=0.33\linewidth,height=0.22\textwidth]{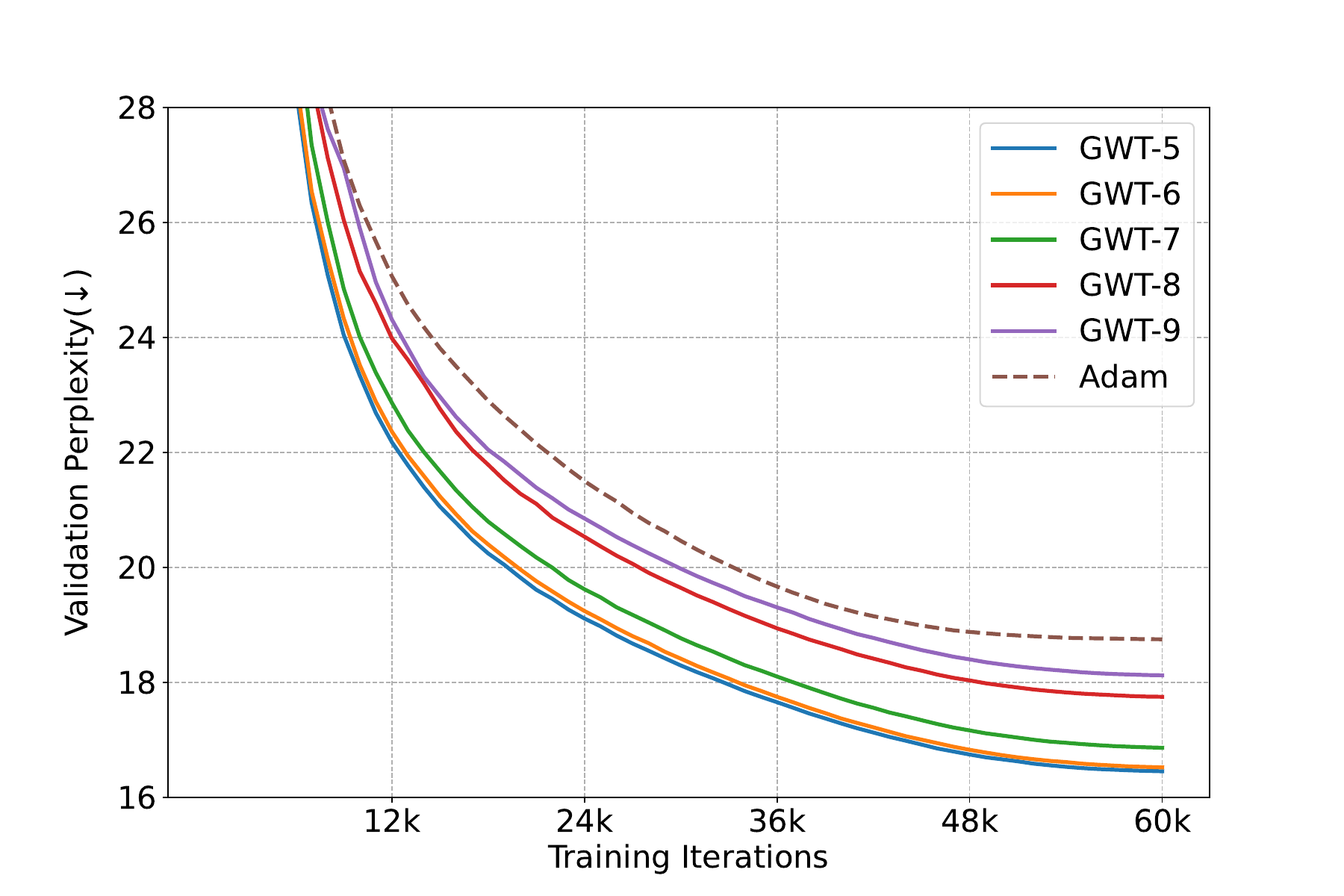}} \\
    \caption{\textbf{Applying GWT to Adam for pre-training LLaMA models with reduced memory usage (higher $l$).} GWT maintains superior performance over Adam under high transform $l$ levels.}\label{fig:learning_curve_level}
\end{figure*}

\textbf{GWT improves optimization efficiency.}
Table~\ref{tab:validation_60m-1b} shows that GWT achieves lower final PPL than full Adam and the matched-compression GaLore variants across the evaluated model scales. Figure~\ref{fig:learning_curve_level} further shows that this behavior remains robust across different decomposition levels. Meanwhile, Table~\ref{tab:llama3b_validation} shows that GWT retains high training throughput: GWT-2 achieves 0.532K tokens per second per GPU on LLaMA-3B, corresponding to a $1.9\times$ speedup over 8-bit Adam~\cite{dettmers20218bit}, about 6\% higher throughput than GaLore-1/4, and performance comparable to the SVD-free APOLLO-1/4. These results show that GWT improves optimization while introducing limited computational overhead.

\begin{table*}[!ht]
    \centering
    \caption{\textbf{Pre-training LLaMA 3B on C4.} We report the validation PPL on different iterations and the token throughput per GPU.}
    \label{tab:llama3b_validation}
    \resizebox{0.95\linewidth}{!}{\begin{tabular}{l|cccccc|cc|c}
    \toprule
      & \textbf{20K} & \textbf{40K} & \textbf{60K} & \textbf{80K} & \textbf{100K}& \textbf{120K}& \textbf{Tokens/s} & \textbf{Time} & \textbf{Mem.}\\
      \midrule
      8bit-Adam & 21.79 & 17.44 & 15.70 & 14.99 & 14.52 & 14.31 & 0.274K & 541.9h &10.7G\\
    \midrule
    GaLore-1/4 & 20.42 & 17.36 & 15.98 & 15.12 & 14.79 & 14.73 & 0.526K & 335.4h & 9.28G \\
    APOLLO-1/4 & 20.78 & 17.19 & 15.40 & 14.40 & 13.90 & 13.75 & 0.541K & 274.4h & 9.28G \\
    \rowcolor{blue4}\textbf{GWT-2} & 19.74 & 16.41 & 14.80 & 13.77 & 13.46 &\textbf{13.21} &0.532K & 279.1h & 8.54G \\
    \midrule
    Tokens (B) & 2.6 & 5.2 & 7.8& 10.4& 13.1 & 15.7 \\
    \bottomrule
    \end{tabular}}
\end{table*}

\textbf{GWT remains efficient under SGD-level memory constraints.}
We assess the performance of GWT at higher decomposition levels, as illustrated in Figure \ref{fig:learning_curve_level}. Notably, even at extreme levels (e.g., $l=7,8,9$), where the memory footprint of the optimizer approaches that of SGD, GWT consistently outperforms full-Adam in terms of validation PPL, demonstrating its robustness and effectiveness in low-memory training regimes.

\begin{figure*}[!th]
    \centering
    \subfloat[LLaMA 60M]{\label{fig:learning_curve_1_adafactor_60m} \includegraphics[width=0.32
    \linewidth]{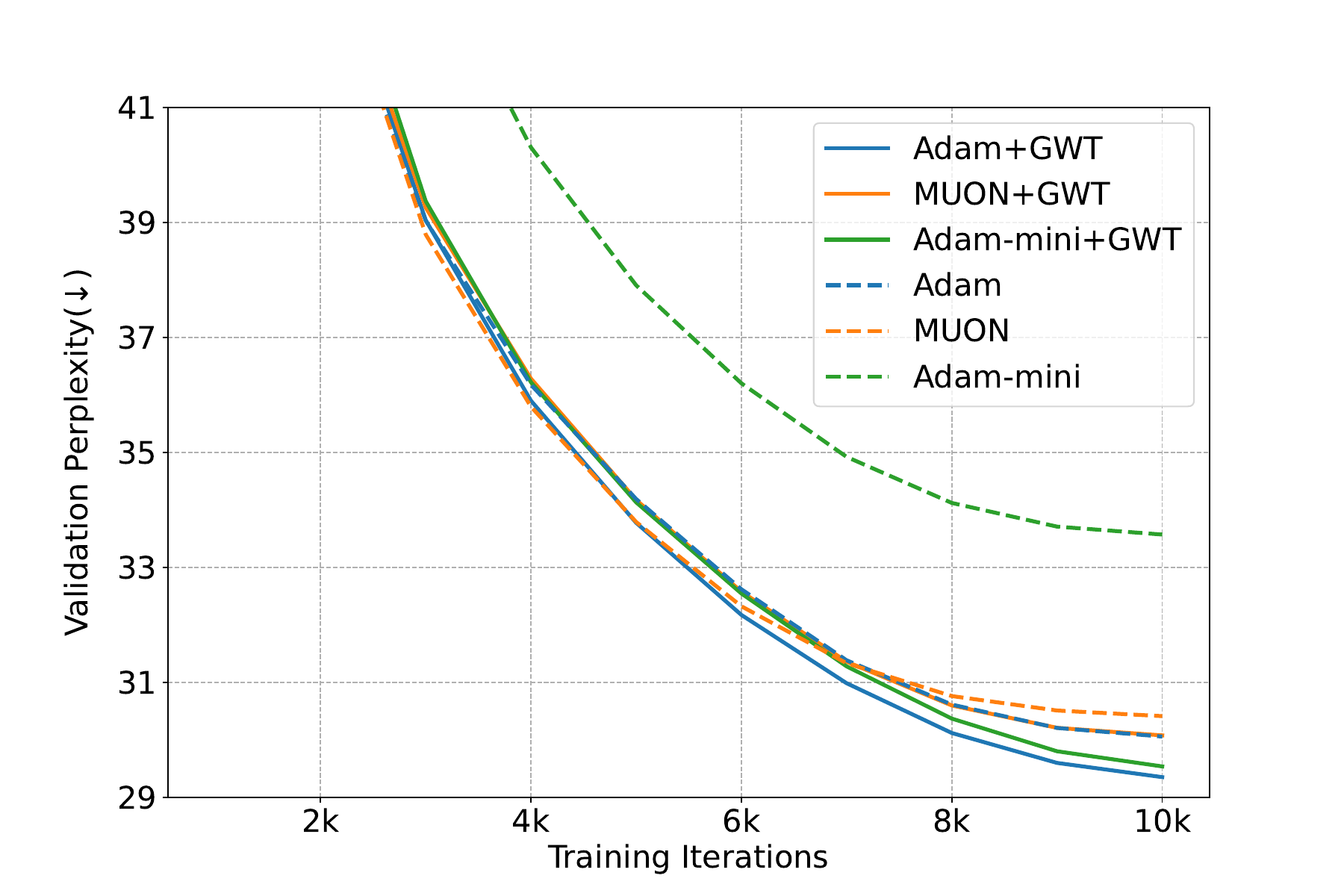}}
    \subfloat[LLaMA 130M]{\label{fig:learning_curve_1_adafactor_130m} \includegraphics[width=0.32\linewidth]{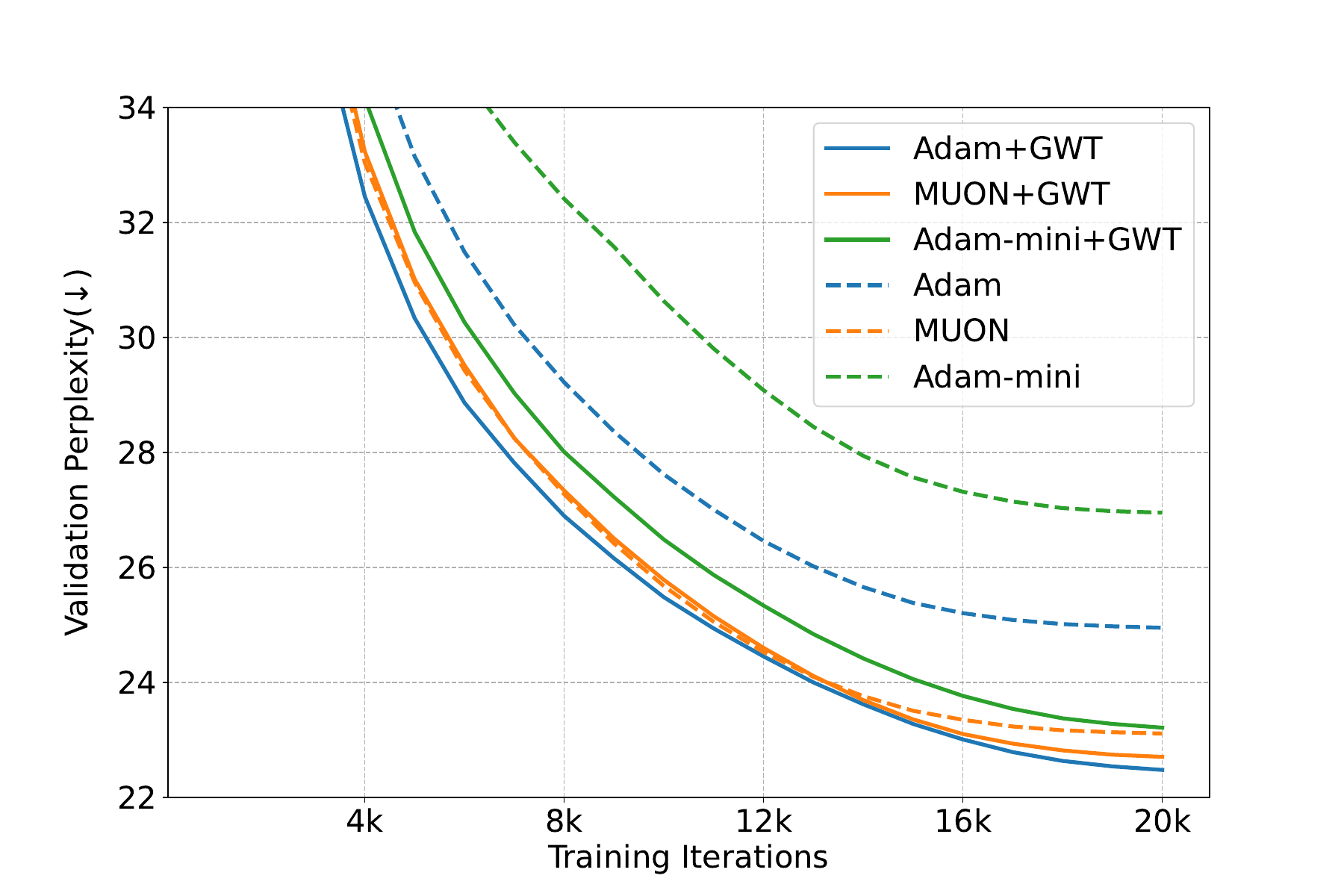}}
    \subfloat[LLaMA 350M]{\label{fig:learning_curve_1_adafactor_350m} \includegraphics[width=0.32
    \linewidth]{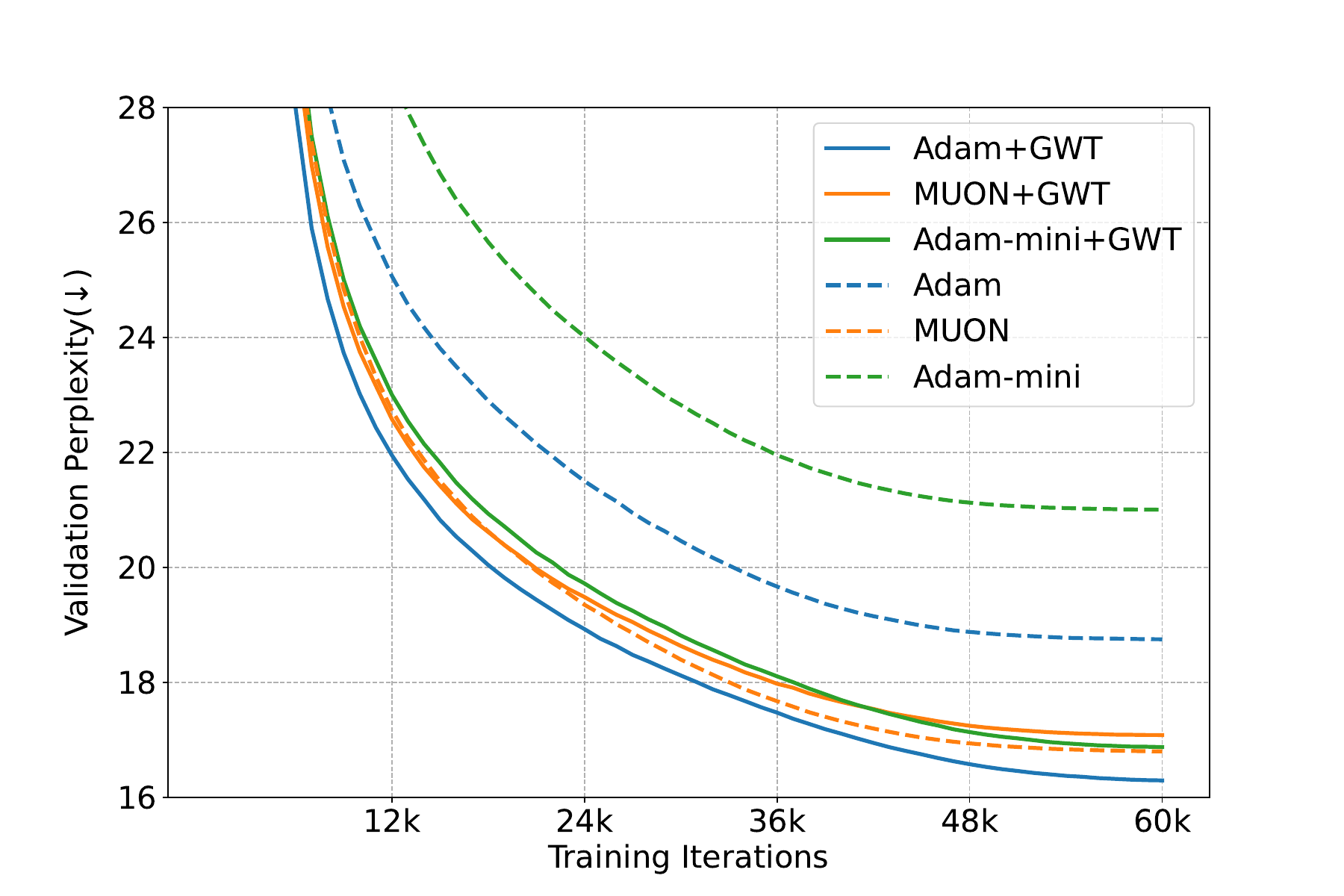}}
    \caption{\textbf{Applying GWT to Adam/Adam-mini/Muon optimizers for pre-training various LLaMA models.} GWT is scalable to optimizers beyond Adam and consistently achieves lower or comparable PPL than full-Adam methods across all tested optimizers.}\label{fig:learning_curve_optimizer}
\end{figure*}

\textbf{GWT is optimizer-agnostic.} We extend our method to the Adam-mini \cite{zhang2024adammini} and Muon \cite{liu2025muon} optimizers and present the corresponding experimental results in Figure \ref{fig:learning_curve_optimizer}. Similar to its performance with Adam, GWT continues to achieve results comparable to or better than the tested methods. This highlights the versatility and effectiveness of our method when integrated with other optimizers, showcasing its broad applicability in memory-efficient optimization.

\textbf{Pre-training with longer sequences.}
We evaluate GWT in handling longer sequences (512, 1024, 2048) while keeping the total number of training tokens per batch constant. Table~\ref{tab:validation_60m-350m_long_sequence} reports the final validation PPL across different methods. In contrast to GaLore, which exhibits noticeable performance degradation as sequence length increases, GWT maintains stable performance, showing only a slight increase in PPL. This resilience highlights GWT as a promising candidate for partial LLM pre-training scenarios, i.e., long-context windows and trillions of training tokens.

\begin{table}[!th]
    \centering
    \caption{\textbf{Final validation PPL on pre-training LLaMA models with different sequence length.} GWT consistently achieves the lowest validation perplexity across all model sizes and sequence lengths, demonstrating robust performance under longer-context training.}
    \label{tab:validation_60m-350m_long_sequence}
    \resizebox{0.95\linewidth}{!}{
    \begin{tabular}{l|ccc|ccc|ccc}
    \toprule
        Models & \multicolumn{3}{c}{\textbf{60M}} & \multicolumn{3}{c}{\textbf{130M}} & \multicolumn{3}{c}{\textbf{350M}} \\
        \midrule
        Sequence length & 512 & 1024 & 2048 & 512 & 1024 & 2048 & 512 & 1024 & 2048 \\
        \midrule
        full-Adam & 34.55 & 37.52 & 39.77 & 25.95 & 28.12 & 32.01 & 19.95 & 22.02 & 24.71 \\
        GaLore-1/4 & 40.25 & 42.02 & 46.31 & 27.19 & 33.97 & 37.29 & 19.92 & 21.73 & 24.10 \\
        APOLLO-1/4 & 32.29 & 34.64 & 37.12 & 24.02 & 25.93 & 28.22 & 17.26 & 18.77 & 19.90 \\
        \midrule
        \rowcolor{blue4}\textbf{GWT-2} & \textbf{30.12} & \textbf{32.55} & \textbf{35.84} & \textbf{23.01} & \textbf{24.89} & \textbf{26.91} & \textbf{16.68} & \textbf{18.21} & \textbf{19.30} \\
         \bottomrule
    \end{tabular}
    }
\end{table}

\subsection{Memory-Efficient Fine-Tuning}

\begin{table}[!t]
    \centering
    \caption{\textbf{Comparison of different fine-tuning methods on MMLU.}
    GWT achieves competitive performance across all three models.}
    \label{tab:fine-tuning_mmlu}
    \resizebox{0.9\linewidth}{!}{
    \begin{tabular}{l|l|cccc|c}
    \toprule
         Models & Methods & \textbf{STEM} & \textbf{Social Sciences} & \textbf{Humanities} & \textbf{Other} & \textbf{Avg.} \\
         \midrule
         \multirow{5}{*}{LLaMA3.2-3B}& Adam & 49.50 & 68.25 & 56.28 & 64.74 & \textbf{59.40} \\
         & LoRA & 49.83 & 67.63 & 56.56 & 63.88 & 59.23  \\
         & GaLore & 49.73 & 68.70 & 55.94 & 64.47 & 59.37 \\
         & APOLLO & 49.17 & 68.18 & 56.32 & 64.71 & 59.32 \\
         \cmidrule{2-7}
         \rowcolor{blue4}\cellcolor{white}& \textbf{GWT} & 49.50 & 68.61 & 56.56 & 64.62 & 59.34  \\
         \midrule
         \multirow{5}{*}{Gemma3-4B}& Adam & 46.59 & 64.74 & 49.71 & 58.91 & 54.46\\
         & LoRA & 45.36 & 63.41 & 48.63 & 58.36 & 53.41  \\
         & GaLore & 46.36 & 64.51 & 49.42 & 58.91 & 54.26 \\
         & APOLLO & 46.49 & 64.28 & 49.39 & 58.67 & 54.17 \\
         \cmidrule{2-7}
         \rowcolor{blue4}\cellcolor{white}& \textbf{GWT} & 46.39 & 64.87 & 49.73 & 59.38 & \textbf{54.56} \\
         \midrule
         \multirow{5}{*}{Qwen2.5-7B}& Adam & - & - & - & - & OOM \\
         & LoRA & 70.44 & 83.23 & 67.80 & 76.90 & 73.85 \\
         & GaLore & 70.21 & 83.49 & 67.52 & 76.87 & 73.76 \\
         & APOLLO & 70.51 & 83.43 & 67.35 & 76.96 & 73.77 \\
         \cmidrule{2-7}
         \rowcolor{blue4}\cellcolor{white}& \textbf{GWT} & 70.38 & 83.88 & 68.06 & 77.14 & \textbf{74.12} \\
         \bottomrule
    \end{tabular}
    }
\end{table}

In this section, we further evaluate the effectiveness of our method by fine-tuning the LLaMA-3.2-3B \cite{grattafiori2024llama3}, Gemma3-4B \cite{team2025gemma3}, and Qwen2.5-7B \cite{yang2024qwen2.5} models, and report their performance on the MMLU benchmark \cite{hendrycks2020measuringmmlu}. For a fair comparison, LoRA, GaLore, and APOLLO are configured with a rank of 8, while GWT is set to a decomposition level of 8 to ensure memory alignment. We report the best accuracy achieved by sweeping the learning rate among [1.0e-6, 2.5e-6, 5.0e-6, 1.0e-5, 2.5e-5, 5.0e-5, 1.0e-4], which ensures a fair evaluation for optimizers with different sensitivities to learning rates. Additionally, to provide a more comprehensive assessment, we include results from fine-tuning the RoBERTa-base model \cite{liu2019roberta} on the GLUE~\cite{wang2018glue} benchmarks.

We report the MMLU results in Table~\ref{tab:fine-tuning_mmlu} and the GLUE results in Table~\ref{tab:fine-tuning_GLUE}. As shown, GWT consistently matches or exceeds the performance of the baselines. These results demonstrate that GWT is not only effective during pre-training but also excels in fine-tuning scenarios, achieving competitive performance across diverse tasks while maintaining memory efficiency.

\begin{table*}[!th]
    \centering

    \caption{{\textbf{Fine-tuning RoBERTa-Base on GLUE benchmarks.} 
    }}
    \label{tab:fine-tuning_GLUE}
    \resizebox{\linewidth}{!}{\begin{tabular}{l|cccccccc|c}
    \toprule
        & \textbf{CoLA} & \textbf{STS-B} & \textbf{MRPC} & \textbf{RTE} & \textbf{SST2} & \textbf{MNLI} & \textbf{QNLI} & \textbf{QQP} & \textbf{Avg.} \\
        \midrule
        Adam & \textbf{63.81}& \textbf{91.26} & 92.25 & 79.06 & \textbf{94.57} & 87.29 & 92.18 & \textbf{92.28} & \textbf{86.58}\\
        \midrule
        GaLore &  61.32 & 91.13 & 92.41 & 77.25& 94.03& 86.77& \textbf{92.56}& 91.77& 85.90\\
        APOLLO & 61.07 & 90.70 & 92.00 & 78.33& 93.57 & 87.21 & 92.27 &91.85 & 85.87\\
        LoRA$^{*}$ \cite{zhao2024galore} & 61.83 & 90.80 &  91.90 &  79.06 & 93.46 & 86.94 &  92.25&  91.22& 85.93\\
        \midrule
        \rowcolor{blue4}\cellcolor{white}GWT  &  62.57 & 91.16 & \textbf{93.26} & \textbf{79.42}  & 94.26 & \textbf{87.37} & 92.53& 91.94 & 86.56\\
         \bottomrule
    \end{tabular}}
\end{table*}

\section{Conclusion}
We introduce GWT, an SVD-free framework that compresses persistent optimizer states in a multilevel Haar domain while retaining all current gradient coefficients in the reconstructed update. Our theoretical and empirical analyses characterize why this design remains effective: the Haar representation captures structured gradient information in a compact approximation band, while the retained detail coefficients preserve complementary instantaneous information and help maintain gradient-aligned updates. Under bounded preconditioning and coarse-momentum coherence, GWT further admits a convergence guarantee. Across language-model pre-training and fine-tuning, GWT reduces memory and improves training throughput while maintaining competitive model quality. Its compatibility with Adam, Adam-mini, and Muon further demonstrates the potential of wavelet-domain state compression as a general mechanism for memory-efficient optimization.

\subsection*{AI Use Statement}

In this work, we used generative AI tools to edit the manuscript for clarity, conciseness, grammar, and English-language presentation. We have not used generative AI tools for other research tasks requiring disclosure, and the remaining disclosure categories are not applicable to this work. We have reviewed all AI-assisted edits and manually verified the final manuscript. We take responsibility for the final content of this work, including text, claims, and artifacts produced with the aid of generative AI.

\subsection*{Ethics statement}

This paper presents work whose goal is to advance the field of Machine Learning. There are many potential societal consequences of our work, none of which we feel must be specifically highlighted here.

\subsection*{Reproducibility statement}

We provide the information required to reproduce the proposed method and experiments throughout the paper and appendix. The assumptions and complete derivations supporting the theoretical results are also provided in the appendix. We will include the implementation and configuration files in the anonymized supplementary material to facilitate reproduction of the reported results.

\bibliography{iclr2027_conference}
\bibliographystyle{iclr2027_conference}

\newpage
\appendix
\onecolumn

\startcontents[appendix]

\phantomsection
\section*{\centering Appendix Contents}
\addcontentsline{toc}{section}{Appendix Contents}

\vspace{0.5em}

\printcontents[appendix]{}{1}{%
\setcounter{tocdepth}{3}%
}

\clearpage

\section{Notation}\label{app:notation}
\begin{table}[!ht]
\centering
\renewcommand{\arraystretch}{1.12}
\begin{tabular}{p{0.24\linewidth}p{0.70\linewidth}}
\toprule
Symbol & Definition \\
\midrule
$t,T$ & Optimization iteration and total number of analyzed iterations; $W_t$ is the parameter matrix before processing $G_t$. \\
$j,l$ & Haar level and total decomposition level, respectively, with $j\in\{1,\ldots,l\}$. \\
$m,n$ & Row and column dimensions of a matrix parameter and its gradient. \\
$F(W)$ & Deterministic objective used in the convergence analysis; $F_*$ is a finite lower bound of $F$. \\
$L$ & Lipschitz constant of $\nabla F$ with respect to the Frobenius norm. \\
$G_t$ & Matrix gradient used at iteration $t$; in Theorem~\ref{thm:gwt_descent}, $G_t=\nabla F(W_t)$. \\
$H_s$ & One-level orthonormal Haar matrix for an even dimension $s$. \\
$\mathcal{H}_l,\mathcal{H}_l^{-1}$ & Level-$l$ Haar transform and its inverse. \\
$A_t^{(j)},D_t^{(j)}$ & Approximation and detail coefficients at Haar level $j$ obtained from $G_t$. \\
$M_t^A,V_t^A$ & Persistent first- and second-moment states for the coarsest approximation coefficients before iteration $t$. \\
$\widehat M_{t+1}^A,\widehat V_{t+1}^A$ & Bias-corrected moment states after incorporating $A_t^{(l)}$. \\
$\mathcal{B}_{j,l}$ & Dyadic shape-alignment operator that expands a coarsest-level tensor to the shape of $D_t^{(j)}$. \\
$S_{t+1}^A,S_{t+1}^{D,j}$ & Positive element-wise preconditioners for the approximation and the level-$j$ detail coefficients. \\
$\widetilde A_t^{(l)},\widetilde D_t^{(j)}$ & Preconditioned approximation and detail update coefficients at iteration $t$. \\
$\Xi_t^A$ & Coarse-momentum error, $\Xi_t^A=\widehat M_{t+1}^A-A_t^{(l)}$. \\
$\omega_{t,q},c_{t,u}$ & Nonnegative temporal weights in Lemma~\ref{lem:momentum_temporal}. \\
$\widetilde G_t$ & Reconstructed GWT update direction before the optional Norm-growth Limiter. \\
$\alpha,\eta$ & Module-wise scale factor and fixed learning rate. \\
$\beta_1,\beta_2,\epsilon$ & Moment decay rates and numerical-stability constant. \\
$P_l(G)$ & Haar low-pass reconstruction obtained by retaining the level-$l$ approximation and setting all detail coefficients to zero. \\
$\Delta G$ & Adjacent-column difference matrix, $(\Delta G)_{:,k}=G_{:,k+1}-G_{:,k}$. \\
$b,\kappa_b$ & Haar block length $b=2^l$ and the discrete Poincar\'e constant $\kappa_b=[2\sin(\pi/(2b))]^{-1}$. \\
$s_{\min},s_{\max}$ & Uniform lower and upper bounds on the entries of the GWT preconditioners. \\
$\rho$ & Coarse-momentum mismatch parameter in Assumption~\ref{assump:gwt_alignment}. \\
$\chi$ & Preconditioner condition ratio, $\chi=s_{\max}/s_{\min}$. \\
$\mu,C$ & Uniform alignment and direction-norm constants in Theorem~\ref{thm:gwt_descent}. \\
$\langle\cdot,\cdot\rangle_F,\|\cdot\|_F$ & Frobenius inner product and norm. \\
\bottomrule
\end{tabular}
\end{table}
\clearpage

\section{Haar Transform and Stabilization Details}\label{app:implementation_details}

\subsection{Eight-Dimensional Haar Example}
For an input vector $x=[x_1,\ldots,x_8]$, the first-level approximation and detail coefficients are
\begin{equation}\label{eq:haar_example}
\begin{aligned}
A^{(1)}&=\frac{1}{\sqrt{2}}[x_1+x_2,\ x_3+x_4,\ x_5+x_6,\ x_7+x_8],\\
D^{(1)}&=\frac{1}{\sqrt{2}}[x_1-x_2,\ x_3-x_4,\ x_5-x_6,\ x_7-x_8].
\end{aligned}
\end{equation}
Applying the transform recursively to $A^{(1)}$ gives
\begin{equation}\nonumber
\begin{aligned}
A^{(2)}&=\frac{1}{2}[x_1+x_2+x_3+x_4,\ x_5+x_6+x_7+x_8],\\
D^{(2)}&=\frac{1}{2}[x_1+x_2-x_3-x_4,\ x_5+x_6-x_7-x_8].
\end{aligned}
\end{equation}
The complete coefficient set $(A^{(2)},D^{(2)},D^{(1)})$ contains the same number of entries as $x$ and reconstructs it exactly. GWT stores persistent optimizer states only for $A^{(2)}$, which contains one quarter of the original entries; the detail coefficients remain temporary.

\begin{figure*}[!th]
    \centering
    \includegraphics[width=0.9\linewidth]{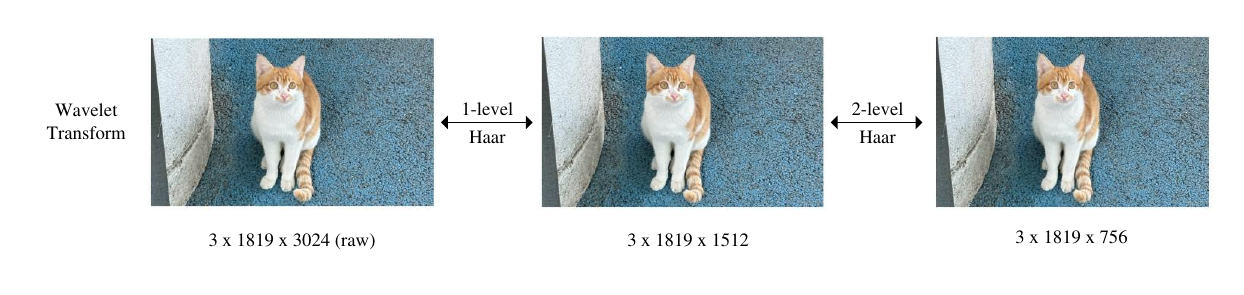}
    \caption{\textbf{Illustration of 2-level Haar approximation coefficients on an image.} The example is included only to visualize the multiresolution decomposition; the empirical gradient-structure diagnostic is discussed separately.}\label{fig:visuliazation_haar}
\end{figure*}

\subsection{Compact GWT Pseudocode}
\begin{algorithm}[!tb]
\caption{Adam with Gradient Wavelet Transform}\label{algo:wavelet_algo}
\begin{algorithmic}
\STATE {\bfseries Input:} Initial weight $W_0$, learning rate $\eta$, decay rates $\beta_1,\beta_2$, scale $\alpha$, level $l$, and $\epsilon>0$
\STATE Initialize $M_0^A\leftarrow 0$ and $V_0^A\leftarrow 0$
\FOR{$t=0,\ldots,T-1$}
    \STATE $G_t\leftarrow \nabla_W f(W_t)$
    \STATE $(A_t^{(l)},D_t^{(l)},\ldots,D_t^{(1)})\leftarrow\mathcal{H}_l(G_t)$
    \STATE Update $M_{t+1}^A,V_{t+1}^A$ using ~\eqref{eq:gwt_state_update}
    \STATE Compute $\widehat M_{t+1}^A,\widehat V_{t+1}^A$ and the preconditioners in ~\eqref{eq:gwt_preconditioners}
    \STATE $\widetilde A_t^{(l)}\leftarrow S_{t+1}^A\odot\widehat M_{t+1}^A$
    \FOR{$j=1,\ldots,l$}
        \STATE $\widetilde D_t^{(j)}\leftarrow S_{t+1}^{D,j}\odot D_t^{(j)}$
    \ENDFOR
    \STATE $\widetilde G_t\leftarrow\alpha\mathcal{H}_l^{-1}(\widetilde A_t^{(l)},\widetilde D_t^{(l)},\ldots,\widetilde D_t^{(1)})$
    \STATE If $t\geq1$, optionally apply the Norm-growth Limiter in ~\eqref{eq:norm_growth_limiter}
    \STATE $W_{t+1}\leftarrow W_t-\eta\widetilde G_t$
\ENDFOR
\STATE \textbf{return} $W_T$
\end{algorithmic}
\end{algorithm}

\subsection{Norm-Growth Limiter}\label{app:norm_growth_limiter}
\begin{figure*}[!th]
    \centering
    \subfloat[LLaMA 60M]{\includegraphics[width=0.48\linewidth,height=0.23\textwidth]{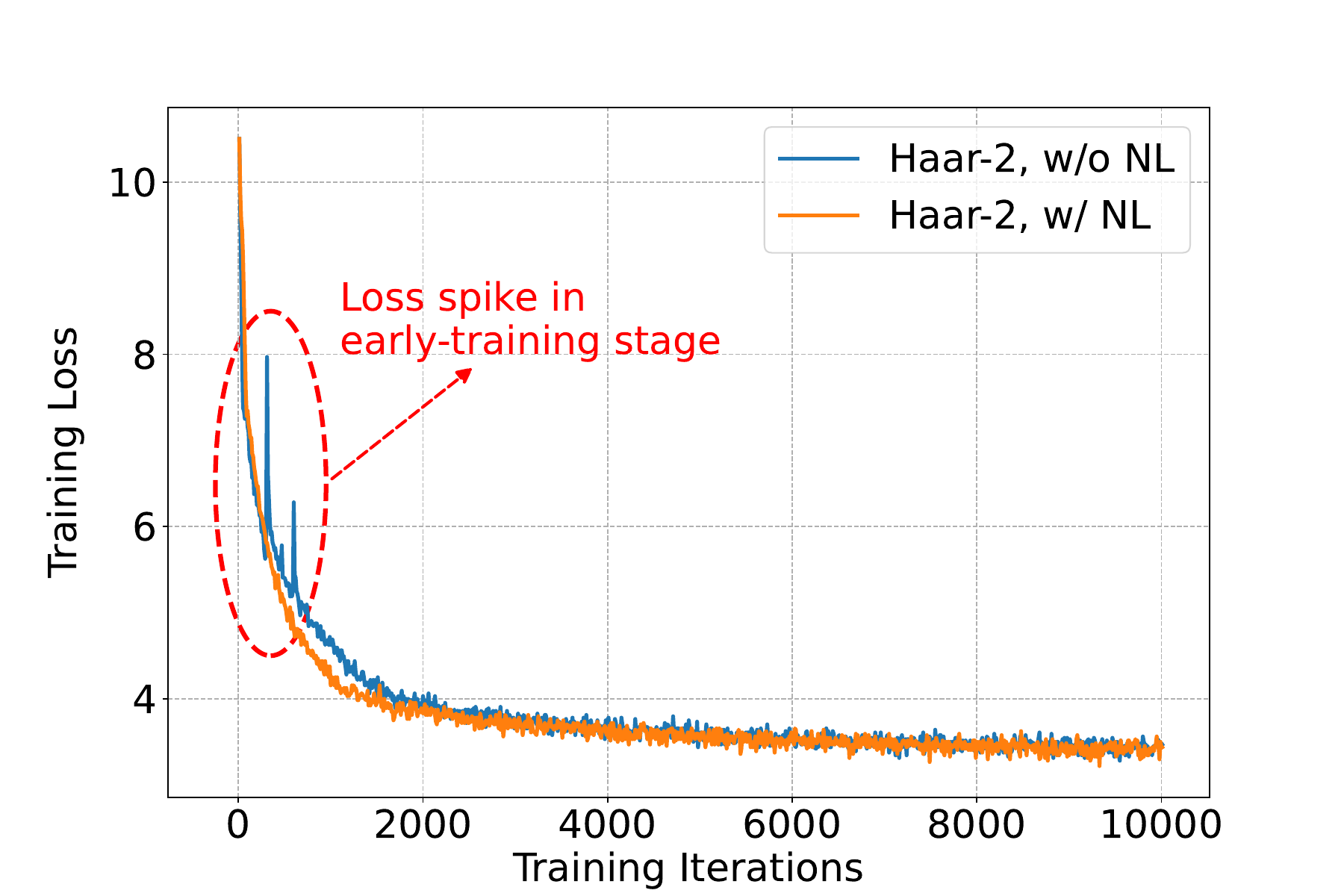}}
    \subfloat[LLaMA 130M]{\includegraphics[width=0.48\linewidth,height=0.23\textwidth]{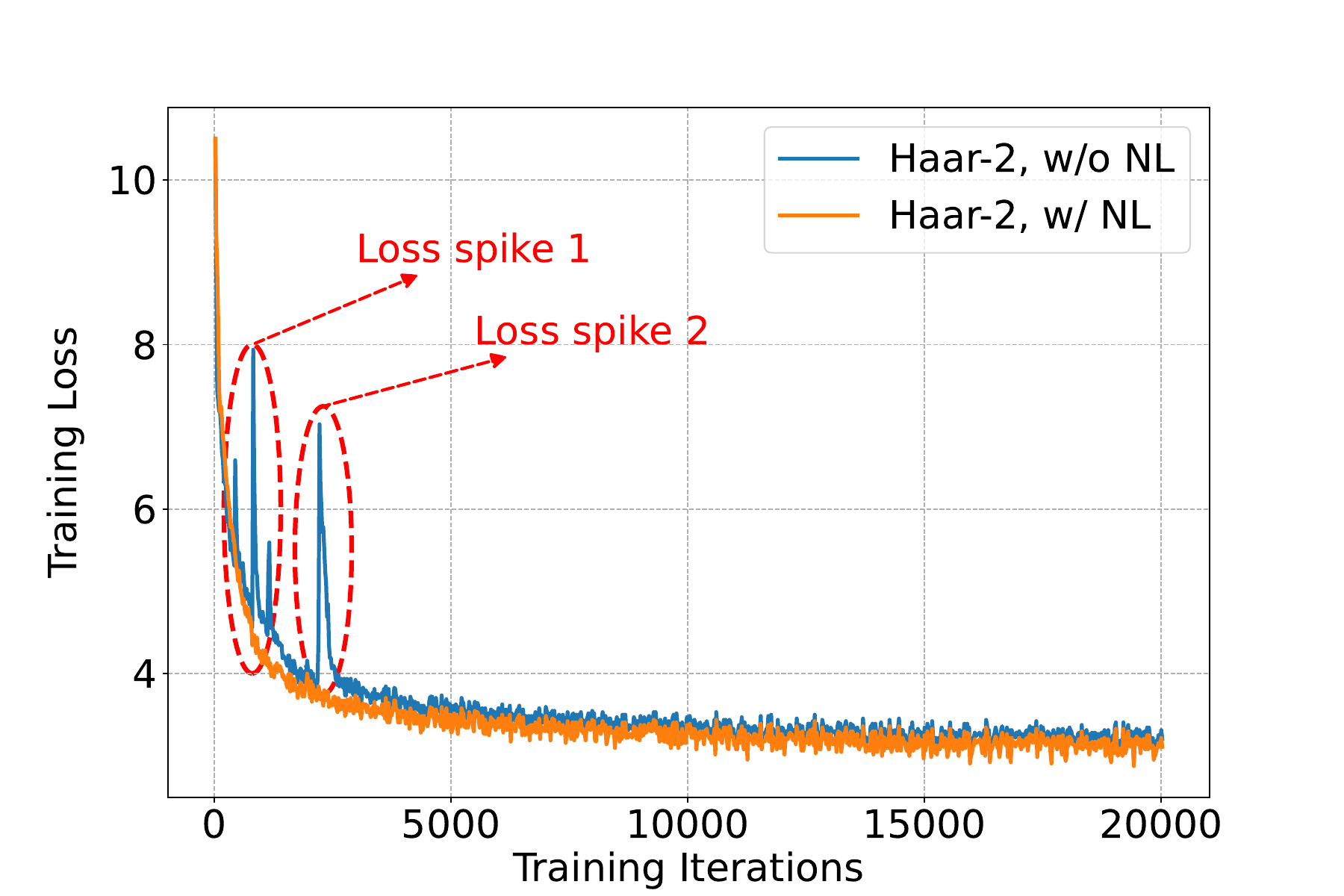}}\\
    \caption{\textbf{Training loss of GWT-2 with and without the Norm-growth Limiter on LLaMA 60M and 130M.}}\label{fig:ablation_nl_llama60m}
\end{figure*}

The Norm-growth Limiter (NL)~\cite{Chen2024FiraCW} controls abrupt changes in the norm of the reconstructed update. For $t\geq1$ and threshold $\gamma>1$, it applies
\begin{equation}\label{eq:norm_growth_limiter}
\text{if}\quad
\frac{\|\widetilde G_t\|_F}{\|\widetilde G_{t-1}\|_F}>\gamma,
\quad\text{then}\quad
\widetilde G_t\leftarrow
\frac{\widetilde G_t}{\|\widetilde G_t\|_F}
\gamma\|\widetilde G_{t-1}\|_F.
\end{equation}
We use $\gamma=1.01$ by default. As shown in Figure~\ref{fig:ablation_nl_llama60m}, NL suppresses the loss spikes observed in early training. Because NL acts after gradient reconstruction and is not specific to the Haar transform, we treat it as an auxiliary stabilization technique.

\section{Proofs for GWT Analysis}\label{app:theory}

This section proves the detail-energy bound, the temporal identity for the approximation-state momentum, and the descent and convergence result stated in Section~\ref{sec:theory}. The proofs follow the same order as the main text: spatial structure of the Haar coefficients, temporal structure of the compressed state, and finally descent and stationary-point convergence of the reconstructed GWT direction.

\subsection{Haar Detail Energy}

We begin with a discrete Poincar\'e inequality for a single block.

\begin{lemma}[Discrete Poincar\'e Inequality]\label{lem:poincare_vector}
Let $x=(x_1,\ldots,x_b)\in\mathbb{R}^b$ and let
\begin{equation}\nonumber
    \bar x=\frac{1}{b}\sum_{k=1}^{b}x_k.
\end{equation}
Define $R_b\in\mathbb{R}^{(b-1)\times b}$ by $(R_b x)_k=x_{k+1}-x_k$. Then
\begin{equation}\label{eq:poincare_vector}
    \|x-\bar x\mathbf{1}\|_2
    \leq \kappa_b\|R_b x\|_2,
    \qquad
    \kappa_b=\frac{1}{2\sin(\pi/(2b))}.
\end{equation}
\end{lemma}

\begin{proof}
Let $L_b=R_b^\top R_b$ be the path-graph Laplacian. Its quadratic form is
\begin{equation}\nonumber
    x^\top L_bx
    =\|R_b x\|_2^2
    =\sum_{k=1}^{b-1}(x_{k+1}-x_k)^2.
\end{equation}
For $h=0,\ldots,b-1$, define $v^{(h)}\in\mathbb{R}^b$ by
\begin{equation}\nonumber
    v_k^{(h)}
    =\cos\left(\left(k-\frac{1}{2}\right)\frac{h\pi}{b}\right),
    \qquad k=1,\ldots,b.
\end{equation}
For an interior index, the cosine addition identity gives
\begin{equation}\nonumber
\begin{aligned}
    (L_bv^{(h)})_k
    &=2v_k^{(h)}-v_{k-1}^{(h)}-v_{k+1}^{(h)}\\
    &=2\left(1-\cos\left(\frac{h\pi}{b}\right)\right)v_k^{(h)}.
\end{aligned}
\end{equation}
The same equality holds at the two endpoints by the first and last rows of $L_b$. Hence $v^{(h)}$ is an eigenvector with eigenvalue
\begin{equation}\nonumber
    \lambda_h=2\left(1-\cos\left(\frac{h\pi}{b}\right)\right),
    \qquad h=0,\ldots,b-1.
\end{equation}
The eigenvalue $\lambda_0=0$ corresponds to the constant vector $\mathbf{1}$, and the smallest positive eigenvalue is
\begin{equation}\nonumber
    \lambda_1=2\left(1-\cos\left(\frac{\pi}{b}\right)\right)
    =4\sin^2\left(\frac{\pi}{2b}\right).
\end{equation}
Set $y=x-\bar x\mathbf{1}$. By construction, $y$ is orthogonal to $\mathbf{1}$ and $R_b y=R_b x$. The Rayleigh quotient on the subspace orthogonal to $\mathbf{1}$ therefore gives
\begin{equation}\nonumber
    y^\top L_b y\geq\lambda_1\|y\|_2^2.
\end{equation}
Using $y^\top L_b y=\|R_b x\|_2^2$ and rearranging yields
\begin{equation}\nonumber
    \|x-\bar x\mathbf{1}\|_2
    \leq\frac{1}{\sqrt{\lambda_1}}\|R_b x\|_2
    =\frac{1}{2\sin(\pi/(2b))}\|R_b x\|_2,
\end{equation}
which proves the claim.
\end{proof}

The vector inequality extends row-wise to a matrix gradient.

\begin{lemma}[Block-Averaging Error]\label{lem:poincare_matrix}
Let $b=2^l$ divide $n$, and let $P_l(G)$ replace every consecutive block of $b$ columns of $G\in\mathbb{R}^{m\times n}$ by its block mean. Then
\begin{equation}\label{eq:poincare_matrix}
    \|G-P_l(G)\|_F
    \leq\kappa_b\|\Delta G\|_F.
\end{equation}
\end{lemma}

\begin{proof}
Fix a row of $G$ and one consecutive block of $b$ columns. Applying Lemma~\ref{lem:poincare_vector} to the entries in this block bounds the squared deviation from the block mean by $\kappa_b^2$ times the sum of squared adjacent differences within the same block. Summing this inequality over all rows and all blocks gives
\begin{equation}\nonumber
\begin{aligned}
    \|G-P_l(G)\|_F^2
    &\leq\kappa_b^2
    \sum_{i=1}^{m}\sum_{q=1}^{n/b}\sum_{k=1}^{b-1}
    \left(G_{i,(q-1)b+k+1}-G_{i,(q-1)b+k}\right)^2.
\end{aligned}
\end{equation}
The right-hand side contains only adjacent differences that remain inside a Haar block. These terms form a subset of all adjacent-column differences in $\Delta G$, and hence
\begin{equation}\nonumber
    \|G-P_l(G)\|_F^2
    \leq\kappa_b^2\|\Delta G\|_F^2.
\end{equation}
Taking square roots proves the result.
\end{proof}

We can now prove Lemma~\ref{lem:detail_energy}.

\begin{proof}[Proof of Lemma~\ref{lem:detail_energy}]
Suppressing the optimization-iteration subscript, the low-pass reconstruction has the coefficient representation
\begin{equation}\nonumber
    P_l(G)=\mathcal{H}_l^{-1}(A^{(l)},0,\ldots,0).
\end{equation}
Subtracting it from the exact reconstruction in ~\eqref{eq:multilevel_haar_inverse} gives
\begin{equation}\nonumber
    G-P_l(G)=\mathcal{H}_l^{-1}(0,D^{(l)},\ldots,D^{(1)}).
\end{equation}
Because the Haar transform is orthogonal, it preserves the Frobenius norm. Therefore,
\begin{equation}\nonumber
    \|G-P_l(G)\|_F^2
    =\sum_{j=1}^{l}\|D^{(j)}\|_F^2.
\end{equation}
Applying Lemma~\ref{lem:poincare_matrix} to the left-hand side yields
\begin{equation}\nonumber
    \sum_{j=1}^{l}\|D^{(j)}\|_F^2
    =\|G-P_l(G)\|_F^2
    \leq\kappa_b^2\|\Delta G\|_F^2,
\end{equation}
which is ~\eqref{eq:detail_energy_bound}.
\end{proof}

\subsection{Temporal Coherence of the Approximation State}

We next prove the representation of the coarse-momentum error used to interpret Assumption~\ref{assump:gwt_alignment}.

\begin{proof}[Proof of Lemma~\ref{lem:momentum_temporal}]
Starting from $M_0^A=0$, repeated substitution into the first recurrence in ~\eqref{eq:gwt_state_update} gives
\begin{equation}\nonumber
    M_{t+1}^A
    =(1-\beta_1)\sum_{q=0}^{t}\beta_1^{t-q}A_q^{(l)}.
\end{equation}
The geometric sum of the scalar coefficients is $1-\beta_1^{t+1}$. Dividing by the same factor in the bias correction therefore yields
\begin{equation}\nonumber
    \widehat M_{t+1}^A
    =\sum_{q=0}^{t}\omega_{t,q}A_q^{(l)},
    \qquad
    \sum_{q=0}^{t}\omega_{t,q}=1.
\end{equation}
Thus, the bias-corrected first moment is a convex combination of the approximation coefficients observed from iteration $0$ through iteration $t$.

Subtracting the current coefficient $A_t^{(l)}$ and using the fact that the weights sum to one gives
\begin{equation}\nonumber
    \Xi_t^A
    =\sum_{q=0}^{t}\omega_{t,q}
    \big(A_q^{(l)}-A_t^{(l)}\big).
\end{equation}
The term with $q=t$ is zero. For every $q<t$, the difference can be expanded as a telescoping sum,
\begin{equation}\nonumber
    A_q^{(l)}-A_t^{(l)}
    =-\sum_{u=q}^{t-1}
    \big(A_{u+1}^{(l)}-A_u^{(l)}\big).
\end{equation}
Substituting this expansion and exchanging the order of the two finite sums gives
\begin{equation}\nonumber
\begin{aligned}
    \Xi_t^A
    &=-\sum_{u=0}^{t-1}
    \left(\sum_{q=0}^{u}\omega_{t,q}\right)
    \big(A_{u+1}^{(l)}-A_u^{(l)}\big).
\end{aligned}
\end{equation}
The cumulative coefficient is
\begin{equation}\nonumber
\begin{aligned}
    \sum_{q=0}^{u}\omega_{t,q}
    &=\frac{1-\beta_1}{1-\beta_1^{t+1}}
    \sum_{q=0}^{u}\beta_1^{t-q}\\
    &=\frac{\beta_1^{t-u}(1-\beta_1^{u+1})}
    {1-\beta_1^{t+1}}
    =c_{t,u}.
\end{aligned}
\end{equation}
This proves the second identity in ~\eqref{eq:temporal_momentum_error}.

Finally, multiply the identity element-wise by $(S_{t+1}^A)^{1/2}$ and apply the triangle inequality. Since every $c_{t,u}$ is nonnegative,
\begin{equation}\nonumber
\begin{aligned}
    \left\|(S_{t+1}^{A})^{1/2}\odot\Xi_t^A\right\|_F
    &\leq\sum_{u=0}^{t-1}c_{t,u}
    \left\|(S_{t+1}^{A})^{1/2}\odot
    \big(A_{u+1}^{(l)}-A_u^{(l)}\big)\right\|_F,
\end{aligned}
\end{equation}
which is ~\eqref{eq:temporal_momentum_bound}. At $t=0$, the bias-corrected first moment equals $A_0^{(l)}$, so $\Xi_0^A=0$ and the empty-sum convention gives the same conclusion.
\end{proof}

\subsection{Deterministic Descent and Convergence}

The proof uses the fact that an orthogonal transform preserves both norms and inner products. For two coefficient tuples
\begin{equation}\nonumber
    \mathcal{C}=(A,D^{(l)},\ldots,D^{(1)}),
    \qquad
    \mathcal{C}'=(A',D'^{(l)},\ldots,D'^{(1)}),
\end{equation}
we have
\begin{equation}\label{eq:haar_inner_product}
\begin{aligned}
    \left\langle\mathcal{H}_l^{-1}(\mathcal{C}),
    \mathcal{H}_l^{-1}(\mathcal{C}')\right\rangle_F
    &=\langle A,A'\rangle_F
    +\sum_{j=1}^{l}\langle D^{(j)},D'^{(j)}\rangle_F,\\
    \left\|\mathcal{H}_l^{-1}(\mathcal{C})\right\|_F^2
    &=\|A\|_F^2+\sum_{j=1}^{l}\|D^{(j)}\|_F^2.
\end{aligned}
\end{equation}
We now prove the alignment and convergence claims. Assumption~\ref{assump:gwt_alignment} provides the same constants $s_{\min}$, $s_{\max}$, and $\rho$ at every iteration considered by the theorem, so the resulting constants $\mu$ and $C$ are uniform in $t$.

\begin{proof}[Proof of Theorem~\ref{thm:gwt_descent}]
For brevity, define the coarse-momentum error
\begin{equation}\nonumber
    \Xi_t^A=\widehat M_{t+1}^A-A_t^{(l)}.
\end{equation}
Using ~\eqref{eq:haar_inner_product} and the definition of the GWT coefficients, the inner product between the current gradient and the reconstructed direction is
\begin{equation}\nonumber
\begin{aligned}
    \langle G_t,\widetilde G_t\rangle_F
    =\alpha\Bigg(&
    \left\langle A_t^{(l)},S_{t+1}^A\odot\widehat M_{t+1}^A\right\rangle_F\\
    &+\sum_{j=1}^{l}
    \left\langle D_t^{(j)},S_{t+1}^{D,j}\odot D_t^{(j)}\right\rangle_F
    \Bigg).
\end{aligned}
\end{equation}
The approximation term can be separated into the contribution of the current coefficient and the momentum error:
\begin{equation}\nonumber
\begin{aligned}
    \left\langle A_t^{(l)},S_{t+1}^A\odot\widehat M_{t+1}^A\right\rangle_F
    &=\left\|(S_{t+1}^A)^{1/2}\odot A_t^{(l)}\right\|_F^2\\
    &\quad+
    \left\langle
    (S_{t+1}^A)^{1/2}\odot A_t^{(l)},
    (S_{t+1}^A)^{1/2}\odot \Xi_t^A
    \right\rangle_F.
\end{aligned}
\end{equation}
By the Cauchy--Schwarz inequality and ~\eqref{eq:coarse_coherence}, the second term is bounded below by
\begin{equation}\nonumber
\begin{aligned}
    &-
    \left\|(S_{t+1}^A)^{1/2}\odot A_t^{(l)}\right\|_F
    \left\|(S_{t+1}^A)^{1/2}\odot \Xi_t^A\right\|_F\\
    &\geq
    -\rho
    \left\|(S_{t+1}^A)^{1/2}\odot A_t^{(l)}\right\|_F^2.
\end{aligned}
\end{equation}
It follows that
\begin{equation}\nonumber
    \left\langle A_t^{(l)},S_{t+1}^A\odot\widehat M_{t+1}^A\right\rangle_F
    \geq
    (1-\rho)
    \left\|(S_{t+1}^A)^{1/2}\odot A_t^{(l)}\right\|_F^2.
\end{equation}
Each detail term is nonnegative and has the exact form
\begin{equation}\nonumber
    \left\langle D_t^{(j)},S_{t+1}^{D,j}\odot D_t^{(j)}\right\rangle_F
    =\left\|(S_{t+1}^{D,j})^{1/2}\odot D_t^{(j)}\right\|_F^2.
\end{equation}
Since every preconditioner entry is at least $s_{\min}$ and $1-\rho\leq1$, combining these inequalities gives
\begin{equation}\nonumber
\begin{aligned}
    \langle G_t,\widetilde G_t\rangle_F
    &\geq\alpha(1-\rho)s_{\min}
    \left(
    \|A_t^{(l)}\|_F^2+
    \sum_{j=1}^{l}\|D_t^{(j)}\|_F^2
    \right)\\
    &=\alpha(1-\rho)s_{\min}\|G_t\|_F^2,
\end{aligned}
\end{equation}
where the last equality follows from ~\eqref{eq:haar_parseval}. This proves the first inequality in ~\eqref{eq:gwt_alignment_bound} with $\mu=\alpha(1-\rho)s_{\min}$.

We next bound the norm of the GWT direction. The lower preconditioner bound and ~\eqref{eq:coarse_coherence} imply
\begin{equation}\nonumber
\begin{aligned}
    \|\Xi_t^A\|_F
    &\leq s_{\min}^{-1/2}
    \left\|(S_{t+1}^A)^{1/2}\odot \Xi_t^A\right\|_F\\
    &\leq \rho s_{\min}^{-1/2}
    \left\|(S_{t+1}^A)^{1/2}\odot A_t^{(l)}\right\|_F\\
    &\leq \rho\sqrt{\frac{s_{\max}}{s_{\min}}}
    \|A_t^{(l)}\|_F
    =\rho\sqrt{\chi}\|A_t^{(l)}\|_F.
\end{aligned}
\end{equation}
Hence the triangle inequality gives
\begin{equation}\nonumber
    \|\widehat M_{t+1}^A\|_F
    \leq(1+\rho\sqrt{\chi})\|A_t^{(l)}\|_F.
\end{equation}
Using ~\eqref{eq:haar_inner_product} and the upper preconditioner bound, we obtain
\begin{equation}\nonumber
\begin{aligned}
    \|\widetilde G_t\|_F^2
    &=\alpha^2\left(
    \|S_{t+1}^A\odot\widehat M_{t+1}^A\|_F^2
    +\sum_{j=1}^{l}\|S_{t+1}^{D,j}\odot D_t^{(j)}\|_F^2
    \right)\\
    &\leq\alpha^2s_{\max}^2\left(
    (1+\rho\sqrt{\chi})^2\|A_t^{(l)}\|_F^2
    +\sum_{j=1}^{l}\|D_t^{(j)}\|_F^2
    \right)\\
    &\leq\alpha^2s_{\max}^2(1+\rho\sqrt{\chi})^2
    \left(
    \|A_t^{(l)}\|_F^2+
    \sum_{j=1}^{l}\|D_t^{(j)}\|_F^2
    \right)\\
    &=C^2\|G_t\|_F^2.
\end{aligned}
\end{equation}
Taking square roots proves the second inequality in ~\eqref{eq:gwt_alignment_bound}.

It remains to show convergence. Since $F$ has an $L$-Lipschitz gradient, the descent lemma applied to $W_{t+1}=W_t-\eta\widetilde G_t$ yields
\begin{equation}\nonumber
    F(W_{t+1})
    \leq F(W_t)
    -\eta\langle\nabla F(W_t),\widetilde G_t\rangle_F
    +\frac{L\eta^2}{2}\|\widetilde G_t\|_F^2.
\end{equation}
Substituting $G_t=\nabla F(W_t)$ and the two bounds established above gives
\begin{equation}\nonumber
    F(W_{t+1})
    \leq F(W_t)
    -\eta\left(\mu-\frac{L\eta C^2}{2}\right)
    \|\nabla F(W_t)\|_F^2.
\end{equation}
The learning-rate condition ensures that the coefficient in parentheses is positive. Summing from $t=0$ to $T-1$ gives
\begin{equation}\nonumber
\begin{aligned}
    \eta\left(\mu-\frac{L\eta C^2}{2}\right)
    \sum_{t=0}^{T-1}\|\nabla F(W_t)\|_F^2
    &\leq F(W_0)-F(W_T)\\
    &\leq F(W_0)-F_*.
\end{aligned}
\end{equation}
Dividing by $T\eta(\mu-L\eta C^2/2)$ proves ~\eqref{eq:gwt_convergence}.
\end{proof}

\section{Experiment Details}\label{sec:exp_detail}

\subsection{Network Architecture}
In this section, we describe the network architectures of the LLaMA (Large Language Model Meta AI) \cite{Touvron2023LLaMAOA} and RoBERTa (Robustly Optimized BERT Approach) \cite{liu2019roberta} models, which complement the experimental details provided in the main text. LLaMA is a family of foundational language models based on transformers \cite{Vaswani2017AttentionIA}. For this paper, we adopt the LLaMA models from previous work \cite{Lialin2023ReLoRAHT}, which use RMSNorm and SwiGLU activations \cite{Touvron2023Llama2O, Touvron2023LLaMAOA}. RoBERTa, on the other hand, is a variant of BERT (Bidirectional Encoder Representations from Transformers) \cite{liu2019roberta}, also built on the transformer architecture. Table \ref{tab:llama_parameter} presents the architectural hyperparameters for the LLaMA and RoBERTa-base models, as well as the pre-training token amounts for LLaMA models across different sizes.

\begin{table*}[!ht]
    \centering
    \setlength{\tabcolsep}{5pt}
    \renewcommand{\arraystretch}{1.1}
    \caption{Architecture hyperparameters of LLaMA for pre-training. Batch size and training data amount are specified in tokens.}
    \label{tab:llama_parameter}
    \begin{tabular}{l|ccccccc}
    \toprule
    Models & Params & Hidden & Intermediate & Heads & Layers & Iteration & Tokens \\
    \midrule
    \multirow{5}{*}{LLaMA} & 60M & 512 & 1376 & 8 & 8 & 10K & 1.3B \\
    & 130M & 768 & 2048 & 12 & 12 & 20K & 2.6B \\
    & 350M & 1024 & 2736 & 16 & 24 & 60K & 7.8B \\
    & 1.3B & 2048 & 5461 & 24 & 32 & 100K & 13.1B \\
    & 3B & 2560 & 6848 & 32& 32 & 120K &15.7B \\
    \midrule
    {RoBERTa}& 125M& 768 & 3072 & 12 & 12 & - & - \\
    \bottomrule
    \end{tabular}
\end{table*}

\subsection{Experiment Hyperparameters}\label{app:hyperparameters}
In this section, we provide detailed information on the hyperparameters and experimental setup used in the experiments discussed in the main text. For the LLaMA pre-training tasks, we follow the experimental configuration used in previous work \cite{zhao2024galore}, which employs the default Adam hyperparameters ($\beta_1 = 0.9, \beta_2 = 0.999, \epsilon = 10^{-6}$), a maximum sequence length of 256 tokens, and gradient accumulation with a batch size of 512. Additionally, we apply a learning rate warm-up for the first 10\% of the iterations and use a cosine annealing schedule with the initial learning rate.

\begin{table*}[!ht]
    \centering
    \setlength{\tabcolsep}{12pt}
    \renewcommand{\arraystretch}{1.1}
    \caption{Hyperparameter $lr,\alpha$, training GPUs, and training time (hours) for different LLaMA models. We report the training time for Adam with our GWT.}
    \label{tab:pre-llama_hyperparameters}
    \begin{tabular}{l|ccccc}
    \toprule
         & 60M &130M & 350M & 1.3B\\
        \midrule
        Full-Adam & 0.005, -& 0.001, - & 0.001, - & 0.0005, - \\
        GWT & 0.01, 0.25 & 0.01, 0.25 & 0.01, 0.25 &  0.01, 0.25  \\
        GaLore & 0.01, 0.25 & 0.01, 0.25 & 0.01, 0.25 & 0.01, 0.25  \\
        APOLLO & 0.01, 1.0 & 0.01, 1.0 & 0.01, 1.0 & 0.01, 1.0 & \\
        \midrule
        GPUs (RTX 3090) & 4 & 4 & 16 & 32 \\
        Time & 0.99h & 4.03h & 11.83 & 40.18h \\
         \bottomrule
    \end{tabular}
\end{table*}

For the GWT method, we perform hyperparameter tuning on the scale parameter $\alpha$ from the set $[0.1,0.15,0.2,0.25]$, selecting the value that yields the best performance across all LLaMA pre-training tasks. We found that our method is not sensitive to scale, and setting $\alpha$ to 0.25 yields good results in most pretraining tasks, which is consistent with the hyperparameters used in previous work \cite{zhao2024galore,zhu2024apollosgdlikememoryadamwlevel_apollo}. For the baseline methods, we use their default hyperparameters: $\alpha = 0.25$ and $lr = 0.01$ for GaLore, and $lr = 0.001$ for Adam (0.005 for 60M and 0.0005 for LLaMA 1.3B and 3B). Consistent with prior work \cite{zhao2024galore}, we enable GWT and GaLore in the attention and MLP modules.

Finally, we summarize the hyperparameter $\alpha$, the GPUs used (RTX 3090), and the training time (in hours) for our method required to reproduce the results shown of Table \ref{tab:validation_60m-1b} in Table \ref{tab:pre-llama_hyperparameters}.

\begin{table*}[!th]
    \centering
    \setlength{\tabcolsep}{10pt}
    \renewcommand{\arraystretch}{1.1}
    \caption{Hyperparameter of different models on the MMLU benchmark.}
    \label{tab:mmul_used_lr}
    \begin{tabular}{l|ccccc}
    \toprule
    \textbf{Hyperparameters} & Adam & LoRA & GaLore & APOLLO & GWT \\
    \midrule
    Rank $r$ & - & 8 & 8 & 8 & - \\
    Level $l$ & - & - & - & - & 8 \\
    Scale $\alpha$ & - & 16 & 2 & 32 & 0.25\\
    $lr$ scheduler & \multicolumn{5}{c}{Cosine}\\
    Warmup steps & \multicolumn{5}{c}{10\%} \\
    Epochs & \multicolumn{5}{c}{3}\\
    Batch size w. accumulation & \multicolumn{5}{c}{32}\\
    Where  & \multicolumn{5}{c}{All}\\
    N-shots & \multicolumn{5}{c}{5} \\
    Cut-off length & \multicolumn{5}{c}{2048} \\
    $lr$ (LLaMA3.2-3B) & 1.0e-6 & 2.5e-5 & 1.0e-5 & 1.0e-6 & 1.0e-5 \\
    $lr$ (Gemma3-4B) & 1.0e-6 & 5.0e-6 & 2.5e-6 & 1.0e-6 & 1.0e-5 \\
    $lr$ (Qwen2.5-7B) & - & 2.5e-6 & 2.5e-6 & 1.0e-6 & 5.0e-5 \\
    \midrule
    \end{tabular}
\end{table*}

For the MMLU fine-tuning tasks, we adopt the implementation from \cite{zheng2024llamafactory} for both the dataset identity and the Alpaca English \cite{alpaca}. For our GWT method, we use a fixed decomposition level of $l = 8$ across all evaluated models. This setting aligns the memory footprint with low-rank-based baselines that employ a rank of 8. The gradient scaling coefficient $\alpha$ is empirically set to $\alpha = 0.25$. For other baselines, we follow the hyperparameter recommendations provided in their original papers: $\alpha = 2$ for GaLore \cite{zhao2024galore}, $\alpha = 32$ for APOLLO \cite{zhu2024apollosgdlikememoryadamwlevel_apollo}, and $\alpha = 16\ (2\times r)$ for LoRA \cite{hu2021lora}. In contrast to pre-training, we apply the memory-efficient fine-tuning technique to all linear layers during this task. We report the best accuracy obtained by sweeping the learning rate over the range [1.0e-6, 2.5e-6, 5.0e-6, 1.0e-5, 2.5e-5, 5.0e-5, 1.0e-4] for all the tested models fine-tuning. A summary of all used hyperparameters is provided in Table~\ref{tab:mmul_used_lr}.

\subsection{Experimental Enviroments}\label{app:enviroment}

We utilize the BF-16 format throughout all pre-training processes to optimize memory usage. All experiments were conducted on a setup of 1 to 8 server nodes, each equipped with four NVIDIA 3090 GPUs and an Intel(R) Xeon(R) Gold 6226R CPU @ 2.90GHz. The Python version used is 3.10, with PyTorch \cite{paszke2017pytorch} version 2.3.1 for pre-training and 2.6.0 for MMLU fine-tuning. For the implementation of GWT, we leverage the PyTorch Wavelet Toolbox \cite{pytorch_wave}.

\section{Additional Experimental Results}

\subsection{Empirical Validation of Coarse-Momentum Coherence}
\label{app:coherence_validation}

The convergence analysis in Section~\ref{sec:theory} relies on the
coarse-momentum coherence condition in Eq.~\eqref{eq:coarse_coherence}.
We empirically examine whether this sufficient condition captures the
behavior of GWT during practical LLM training. Specifically, we train
LLaMA-60M and LLaMA-130M for 10K optimization steps using GWT-2 and
GWT-3. All other settings follow the corresponding pre-training
experiments in the main text, including the optimizer hyperparameters,
batch size, sequence length, warm-up, and learning-rate schedule. During
training, we periodically record gradients from representative layers and
all GWT-optimized attention and MLP linear matrices within these layers.

For each monitored matrix gradient $G_t$, we compute the empirical
coherence ratio
\begin{equation}
    \rho_t^{\mathrm{emp}}
    =
    \frac{
        \left\|
        (S_{t+1}^{A})^{1/2}\odot
        \left(\widehat M_{t+1}^{A}-A_t^{(l)}\right)
        \right\|_F
    }{
        \left\|
        (S_{t+1}^{A})^{1/2}\odot A_t^{(l)}
        \right\|_F
    }.
    \label{eq:empirical_coherence_ratio}
\end{equation}
Here, $A_t^{(l)}$ is the current coarsest Haar approximation coefficient,
$\widehat M_{t+1}^{A}$ is its bias-corrected first-moment estimate, and
$S_{t+1}^{A}$ is the corresponding preconditioner. Thus,
$\rho_t^{\mathrm{emp}}$ measures the magnitude of the preconditioned
coarse-momentum error relative to the current approximation coefficient.
The condition $\rho_t^{\mathrm{emp}}<1$ directly corresponds to the
norm-based sufficient coherence condition in
Assumption~\ref{assump:gwt_alignment}. Each observation in the following
statistics corresponds to one monitored matrix at one sampled optimization
step.

\begin{figure*}[!th]
    \centering
    \begin{subfigure}[b]{0.46\linewidth}
        \centering
        \includegraphics[width=\linewidth]
        {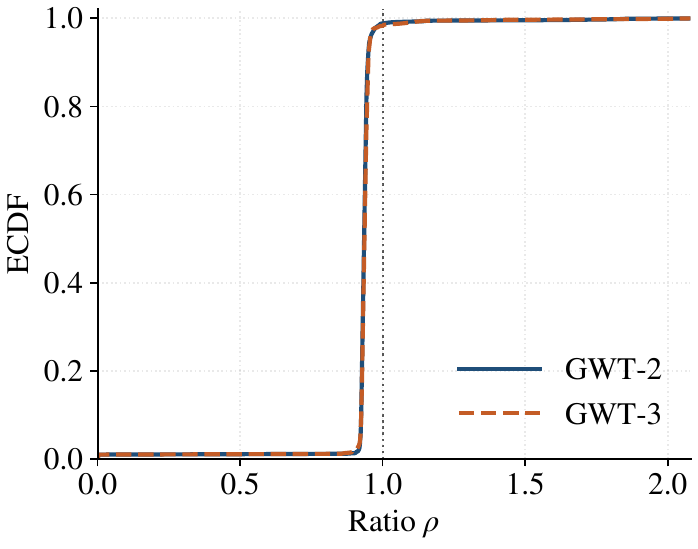}
        \caption{LLaMA-60M}
    \end{subfigure}
    \hfill
    \begin{subfigure}[b]{0.46\linewidth}
        \centering
        \includegraphics[width=\linewidth]
        {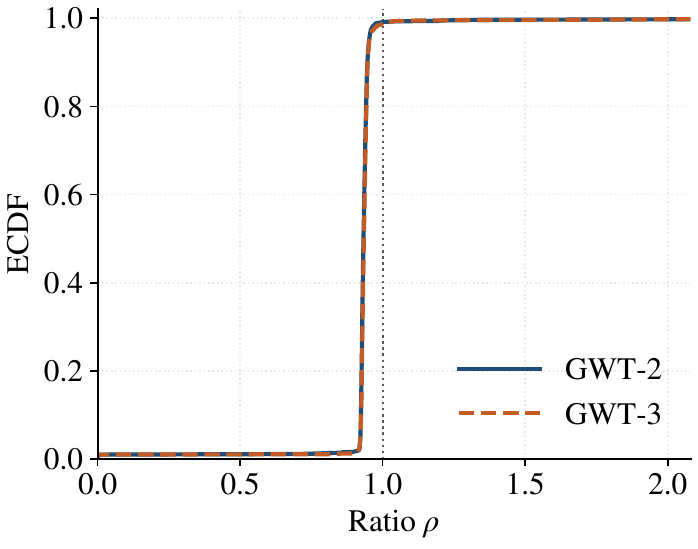}
        \caption{LLaMA-130M}
    \end{subfigure}
    \caption{\textbf{Empirical distribution of coarse-momentum coherence.}
    The horizontal axis denotes the empirical coherence ratio
    $\rho_t^{\mathrm{emp}}$, and the vertical axis is its ECDF, i.e., the
    fraction of sampled matrix--step pairs with coherence ratio no larger
    than the corresponding value. The vertical reference at $\rho=1$
    denotes the sufficient coherence threshold in
    Assumption~\ref{assump:gwt_alignment}. Both GWT levels exhibit closely
    matched distributions across LLaMA-60M and LLaMA-130M.}
    \label{fig:rho_distribution}
\end{figure*}

Figure~\ref{fig:rho_distribution} first reports the empirical cumulative
distribution function (ECDF) of $\rho_t^{\mathrm{emp}}$ over all sampled
matrix--step pairs. For a threshold $x$, the ECDF gives the fraction of
samples satisfying $\rho_t^{\mathrm{emp}}\leq x$; hence, its value around
$x=1$ directly reflects how frequently the sufficient coherence condition
is satisfied. The distributions are strongly concentrated below $1$ for
both model sizes and both decomposition levels. As summarized in
Table~\ref{tab:coherence_validation}, the median ratios lie in the narrow
range $0.933$--$0.936$, while the 95th percentiles remain between
$0.952$ and $0.955$. Overall, $98.24\%$--$99.05\%$ of all sampled
matrix--step pairs satisfy $\rho_t^{\mathrm{emp}}<1$.

\begin{figure*}[!th]
    \centering
    \begin{subfigure}[b]{0.46\linewidth}
        \centering
        \includegraphics[width=\linewidth]
        {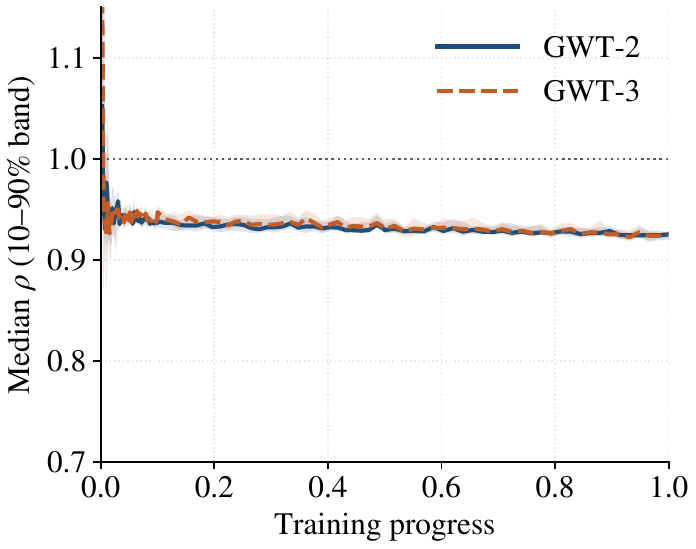}
        \caption{LLaMA-60M}
    \end{subfigure}
    \hfill
    \begin{subfigure}[b]{0.46\linewidth}
        \centering
        \includegraphics[width=\linewidth]
        {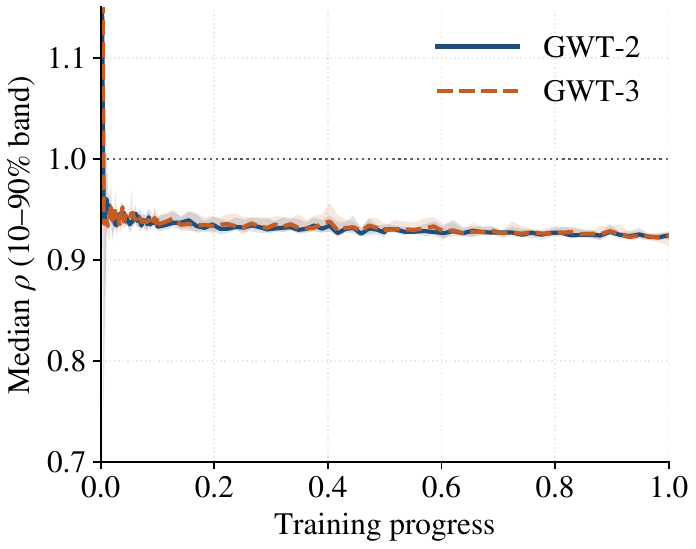}
        \caption{LLaMA-130M}
    \end{subfigure}
    \caption{\textbf{Evolution of coarse-momentum coherence during
    pre-training.}
    At each sampled optimization step, the solid curve denotes the median
    $\rho_t^{\mathrm{emp}}$ across all monitored matrices, and the shaded
    region denotes the corresponding P10--P90 range across matrices.
    Hence, the shaded band measures cross-layer/module dispersion at the
    same step rather than variation across random seeds or temporal
    smoothing. The horizontal reference at $\rho=1$ marks the sufficient
    coherence threshold. The few violations are concentrated in the very
    early stage of training.}
    \label{fig:rho_training}
\end{figure*}

Figure~\ref{fig:rho_training} further preserves the temporal dimension
and examines how the coherence ratio evolves throughout the 10K-step
training trajectory. At each sampled optimization step, we collect
$\rho_t^{\mathrm{emp}}$ from all monitored matrices. The solid curve
shows the median across matrices, while the shaded region spans the
10th--90th percentiles (P10--P90) of the same cross-matrix distribution.
The shaded band therefore characterizes variation across different layers
and modules at a fixed optimization step; it is neither a confidence
interval across random seeds nor a temporal moving window.

The median and the central P10--P90 range remain below $\rho=1$ over
nearly the entire training trajectory. The rare violations are strongly
localized in time and occur almost exclusively during the first 50
optimization steps, when the optimizer states are still in their initial
transient stage. These early samples are retained in all aggregate
statistics. After this short initial period, the monitored matrices
maintain stable coarse-momentum coherence throughout the remainder of
training.

\begin{figure*}[!th]
    \centering
    \begin{subfigure}[b]{0.46\linewidth}
        \centering
        \includegraphics[width=\linewidth]
        {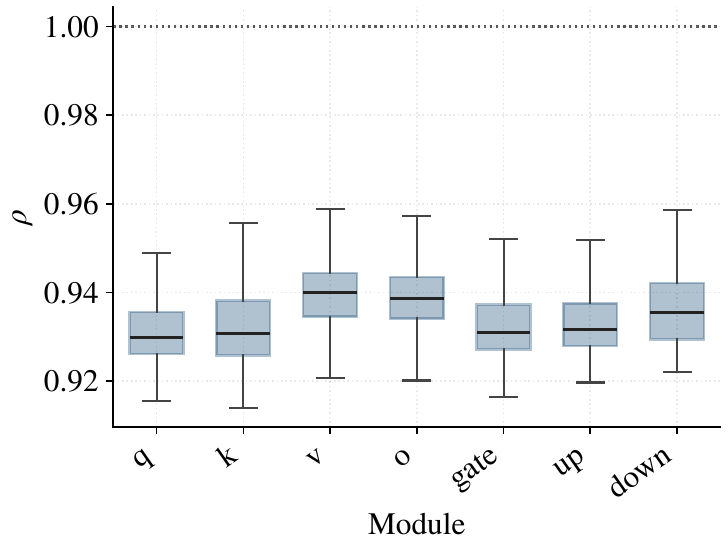}
        \caption{LLaMA-60M, GWT-2}
    \end{subfigure}
    \hfill
    \begin{subfigure}[b]{0.46\linewidth}
        \centering
        \includegraphics[width=\linewidth]
        {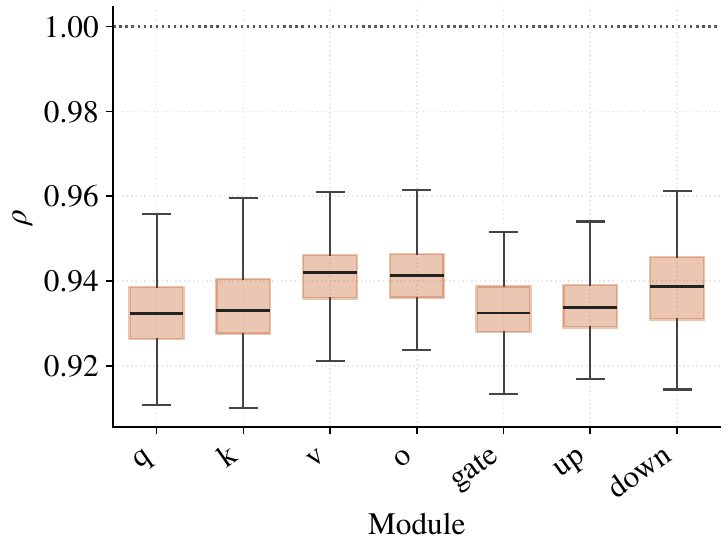}
        \caption{LLaMA-60M, GWT-3}
    \end{subfigure}

    \vspace{2mm}

    \begin{subfigure}[b]{0.46\linewidth}
        \centering
        \includegraphics[width=\linewidth]
        {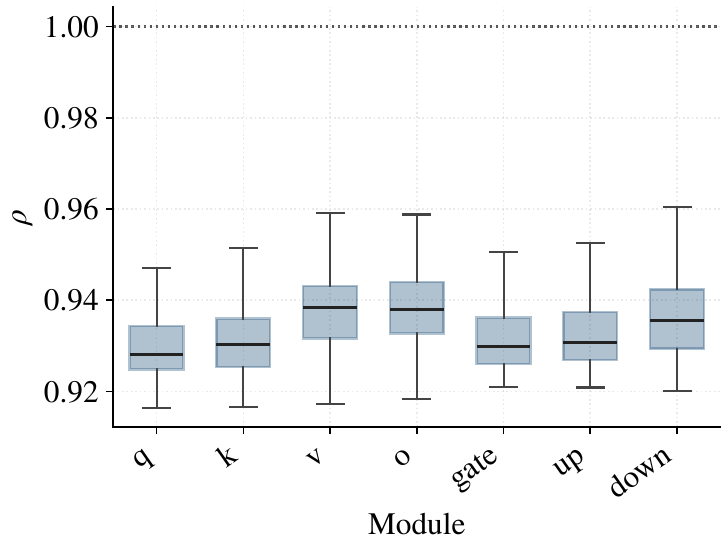}
        \caption{LLaMA-130M, GWT-2}
    \end{subfigure}
    \hfill
    \begin{subfigure}[b]{0.46\linewidth}
        \centering
        \includegraphics[width=\linewidth]
        {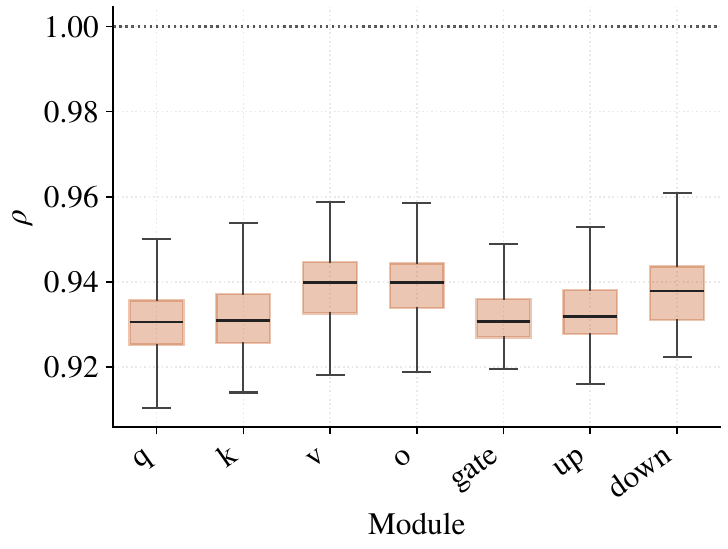}
        \caption{LLaMA-130M, GWT-3}
    \end{subfigure}
    \caption{\textbf{Module-wise coarse-momentum coherence.}
    Boxplots show the empirical distributions of
    $\rho_t^{\mathrm{emp}}$ for different attention and MLP linear
    modules. The horizontal reference at $\rho=1$ denotes the sufficient
    coherence threshold. Across both model sizes and decomposition levels,
    the central distributions of the monitored modules remain consistently
    below this threshold, showing that coarse-momentum coherence is not
    confined to a particular module type.}
    \label{fig:module_rho}
\end{figure*}

We further examine whether the aggregate coherence behavior is dominated
by particular matrix types. Figure~\ref{fig:module_rho} reports the
distribution of $\rho_t^{\mathrm{emp}}$ separately for the attention and
MLP linear modules. Across both LLaMA-60M and LLaMA-130M, and under both
GWT-2 and GWT-3, the medians and central box ranges of essentially all
module types remain clearly below $\rho=1$. Moreover, the module-wise
distributions are broadly similar to one another. This indicates that the
high overall satisfaction rate is not driven by a small subset of layers
or matrix types, but is a common property across the monitored Transformer
modules.

The empirical behavior is also highly consistent when increasing the
model size from LLaMA-60M to LLaMA-130M. Both models have median
$\rho$ values of approximately $0.93$, P95 values of approximately
$0.95$, and satisfaction rates above $98\%$ for both GWT-2 and GWT-3.
Together with the similar module-wise distributions, this suggests that
the observed coarse-momentum coherence is not specific to the smallest
model and remains stable under the model-scale increase considered here.

\begin{table}[!ht]
    \centering
    \caption{\textbf{Empirical validation of coarse-momentum coherence.}
    Statistics are computed over all sampled matrix--step pairs, including
    the initial transient stage. The last column reports the fraction
    satisfying the sufficient condition $\rho_t^{\mathrm{emp}}<1$.}
    \label{tab:coherence_validation}
    \begin{tabular}{l|c|cccc}
        \toprule
        Model & Setting & Median $\rho$ & P95 $\rho$ & Max $\rho$
        & $P(\rho<1)$ \\
        \midrule
        \multirow{2}{*}{LLaMA-60M}
        & GWT-2 & 0.934 & 0.955 & 2.219 & 98.76\% \\
        & GWT-3 & 0.936 & 0.955 & 2.129 & 98.24\% \\
        \midrule
        \multirow{2}{*}{LLaMA-130M}
        & GWT-2 & 0.933 & 0.952 & 4.283 & 99.05\% \\
        & GWT-3 & 0.934 & 0.953 & 4.250 & 98.71\% \\
        \bottomrule
    \end{tabular}
\end{table}

Taken together, the ECDF, temporal evolution, and module-wise
distributions consistently support the empirical relevance of the
coarse-momentum coherence condition. We emphasize that
$\rho_t^{\mathrm{emp}}<1$ is a sufficient rather than necessary
condition. In particular, the small number of early samples exceeding
the threshold does not imply that the complete reconstructed GWT
direction becomes negatively aligned with the instantaneous gradient.
We examine this property next.

\subsection{Understanding GWT: Energy Concentration and Directional Alignment}
\label{app:gwt_understanding}

We next investigate why GWT remains effective despite maintaining
persistent optimizer states only for the coarsest approximation
coefficients. We examine three complementary properties: whether the
compact approximation band captures a disproportionate fraction of the
instantaneous gradient energy, whether this concentration depends on the
original local column structure of the gradient matrix, and whether the
complete reconstructed GWT direction remains aligned with the
instantaneous gradient.

\paragraph{Energy concentration in the approximation band.}
For each sampled gradient, we define the approximation-energy ratio as
\begin{equation}
    R_{A,t}
    =
    \frac{\|A_t^{(l)}\|_F^2}{\|G_t\|_F^2}.
    \label{eq:empirical_approx_energy}
\end{equation}
Since the Haar transform is orthogonal, $R_{A,t}$ measures the exact
fraction of instantaneous gradient energy contained in the coarsest
approximation band, with the complementary fraction carried by the detail
bands. In contrast, the approximation band contains only a fraction
$2^{-l}$ of the transformed coefficients, which is also the fraction for
which GWT maintains persistent optimizer states. We therefore define the
energy enrichment as the median approximation-energy ratio divided by
$2^{-l}$.

\begin{figure*}[!th]
    \centering
    \begin{subfigure}[b]{0.46\linewidth}
        \centering
        \includegraphics[width=\linewidth]
        {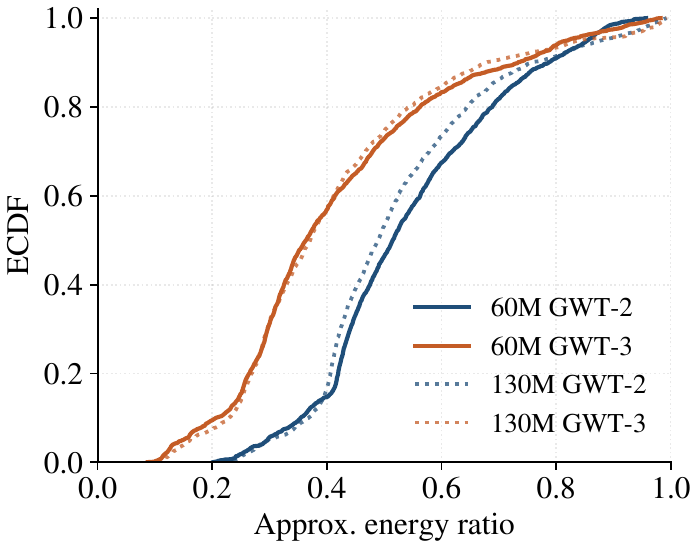}
        \caption{ECDF of approximation-energy ratio}
    \end{subfigure}
    \hfill
    \begin{subfigure}[b]{0.46\linewidth}
        \centering
        \includegraphics[width=\linewidth]
        {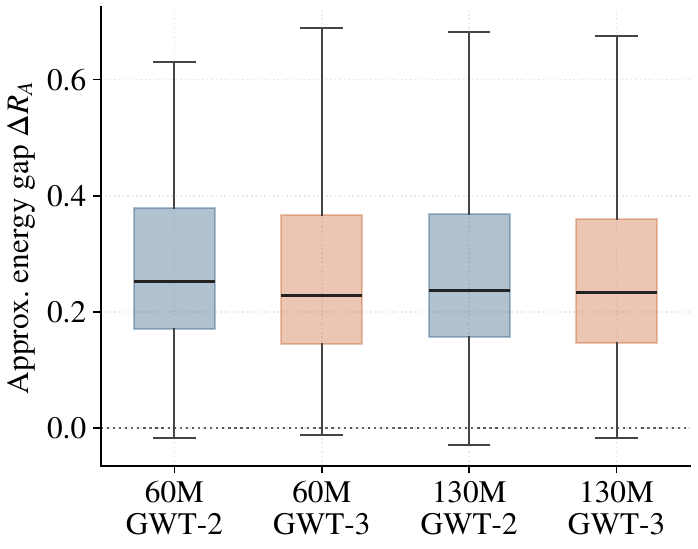}
        \caption{Approximation-energy gap under column permutation}
    \end{subfigure}
    \caption{\textbf{Gradient energy structure in the Haar domain.}
    Left: ECDF of the approximation-energy ratio
    $R_{A,t}=\|A_t^{(l)}\|_F^2/\|G_t\|_F^2$. The horizontal axis denotes
    the fraction of instantaneous gradient energy contained in the
    coarsest approximation band, and the vertical axis denotes its
    empirical cumulative distribution. Right: boxplots of the permutation
    energy gap $\Delta R_{A,t}$, defined as the approximation-energy ratio
    under the original column ordering minus its mean over random column
    permutations. Positive values indicate that the original ordering
    concentrates more gradient energy in the Haar approximation band than
    the permutation control.}
    \label{fig:gwt_energy_structure}
\end{figure*}

The left panel of Figure~\ref{fig:gwt_energy_structure} reports the ECDF
of $R_{A,t}$ over all sampled matrix--step pairs. The horizontal axis is
the approximation-energy ratio, while the vertical axis gives the
cumulative fraction of samples with approximation energy no larger than
the corresponding value. The distributions show that the compact
approximation band captures a substantial fraction of the instantaneous
gradient energy. As summarized in Table~\ref{tab:gwt_energy_structure},
GWT-2 maintains persistent states for only $25\%$ of the transformed
coefficients, yet the median approximation-energy ratios reach $51.5\%$
and $48.9\%$ on LLaMA-60M and LLaMA-130M, respectively. For GWT-3,
only $12.5\%$ of the coefficients retain persistent states, while the
corresponding median energy ratios remain $36.4\%$ and $37.1\%$.
These values correspond to enrichment factors of
$1.96\times$--$2.96\times$ across the four settings.

\paragraph{Dependence on the original column structure.}
The detail-energy analysis in Section~\ref{sec:theory} relates the
approximation-band energy to local variation between neighboring gradient
columns. We therefore perform a column-permutation control to determine
whether the observed concentration depends on the original ordering of
the matrix columns.

For each gradient selected for the permutation diagnostic, we compute the
approximation-energy ratio $R_{A,t}^{\mathrm{orig}}$ under its original
column ordering. We then independently permute the matrix columns several
times, apply the same Haar transform to each permutation, and average the
resulting approximation-energy ratios. We define the permutation energy
gap as
\begin{equation}
    \Delta R_{A,t}
    =
    R_{A,t}^{\mathrm{orig}}
    -
    \frac{1}{K_{\mathrm{perm}}}
    \sum_{k=1}^{K_{\mathrm{perm}}}
    R_{A,t}^{\mathrm{perm},k},
    \label{eq:permutation_energy_gap}
\end{equation}
where $K_{\mathrm{perm}}$ denotes the number of random column
permutations. A positive $\Delta R_{A,t}$ therefore means that the
original column ordering places more gradient energy in the Haar
approximation band than the average random permutation of the same
gradient matrix.

The right panel of Figure~\ref{fig:gwt_energy_structure} reports boxplots
of $\Delta R_{A,t}$ for the four model/level configurations, using
approximately 210 permutation-diagnostic samples per configuration. The
distributions lie predominantly above zero, showing that the original
column ordering systematically yields higher approximation-band energy
than the randomly permuted controls. This behavior is also consistent
across the two model sizes and both GWT levels. Quantitatively,
Table~\ref{tab:gwt_energy_structure} reports
$P(\Delta R_A>0)$, which ranges from $94.8\%$ to $98.1\%$. Since random
column permutation preserves the entries and total Frobenius energy of
each gradient while disrupting adjacency relationships, the positive
energy gaps indicate that Haar exploits non-random local structure in the
original gradient matrices.

\begin{table}[!ht]
    \centering
    \caption{\textbf{Energy concentration in the Haar approximation band.}
    State Frac.\ is the fraction $2^{-l}$ of transformed coefficients for
    which GWT maintains persistent optimizer states. Approx.\ Energy is the
    median approximation-energy ratio, and Enrichment is its ratio to the
    state fraction. The final column reports the fraction of
    permutation-diagnostic samples with a positive energy gap
    $\Delta R_A>0$.}
    \label{tab:gwt_energy_structure}
    \begin{tabular}{l|c|cccc}
        \toprule
        Model & Setting
        & State Frac.
        & Approx. Energy
        & Enrichment
        & $P(\Delta R_A>0)$ \\
        \midrule
        \multirow{2}{*}{LLaMA-60M}
        & GWT-2 & 25.0\% & 51.5\% & $2.06\times$ & 96.7\% \\
        & GWT-3 & 12.5\% & 36.4\% & $2.91\times$ & 97.6\% \\
        \midrule
        \multirow{2}{*}{LLaMA-130M}
        & GWT-2 & 25.0\% & 48.9\% & $1.96\times$ & 98.1\% \\
        & GWT-3 & 12.5\% & 37.1\% & $2.96\times$ & 94.8\% \\
        \bottomrule
    \end{tabular}
\end{table}

\paragraph{Directional alignment of the reconstructed update.}
Energy concentration alone does not guarantee that state compression
produces a useful optimization direction. We therefore directly measure
the cosine similarity between the instantaneous gradient $G_t$ and the
reconstructed GWT direction $\widetilde G_t$ before the optional
Norm-growth Limiter:
\begin{equation}
    \operatorname{CosAlign}_t
    =
    \frac{
        \langle G_t,\widetilde G_t\rangle_F
    }{
        \|G_t\|_F\|\widetilde G_t\|_F
    }.
    \label{eq:empirical_cosine_alignment}
\end{equation}
A positive value indicates that the reconstructed direction has a positive
inner product with the current gradient.

We additionally decompose the gradient--update inner product into its
approximation- and detail-band contributions. The approximation band
contains the temporally accumulated first-moment state, whereas each
current detail band contributes a nonnegative preconditioned term. We
report the normalized detail contribution to the complete
gradient--update inner product; this quantity can exceed one when the
approximation-band contribution is negative.

As shown in Table~\ref{tab:gwt_direction_alignment}, the reconstructed
GWT direction is positively aligned with the instantaneous gradient for
every sampled matrix--step pair across both model sizes and decomposition
levels. The median cosine similarities range from $0.649$ to $0.724$,
and even the minimum observed similarities remain positive,
between $0.195$ and $0.217$. This includes the rare samples for which
$\rho_t^{\mathrm{emp}}\geq1$, empirically confirming that the coherence
condition is sufficient rather than necessary for positive directional
alignment.

The retained detail coefficients also make a substantial contribution to
the complete update. Their median normalized contribution ranges from
$93.7\%$ to $96.7\%$ across the four settings. Hence, while persistent
optimizer states are compressed into the approximation band, the current
detail coefficients preserve the complementary gradient information and
continue to provide a positive contribution to the reconstructed
direction.

\begin{table}[!ht]
    \centering
    \caption{\textbf{Directional alignment of the reconstructed GWT
    update.}
    Cosine similarity is measured between the instantaneous gradient
    $G_t$ and the reconstructed direction $\widetilde G_t$ before the
    optional Norm-growth Limiter. Detail Contribution denotes the median
    normalized contribution of the retained current detail coefficients
    to the gradient--update inner product.}
    \label{tab:gwt_direction_alignment}
    \begin{tabular}{l|c|cccc}
        \toprule
        Model & Setting
        & Median Cos.
        & Min Cos.
        & $P(\mathrm{Cos}>0)$
        & Detail Contribution \\
        \midrule
        \multirow{2}{*}{LLaMA-60M}
        & GWT-2 & 0.649 & 0.215 & 100\% & 93.7\% \\
        & GWT-3 & 0.717 & 0.203 & 100\% & 96.6\% \\
        \midrule
        \multirow{2}{*}{LLaMA-130M}
        & GWT-2 & 0.653 & 0.217 & 100\% & 93.9\% \\
        & GWT-3 & 0.724 & 0.195 & 100\% & 96.7\% \\
        \bottomrule
    \end{tabular}
\end{table}

Taken together, these diagnostics provide complementary evidence for the
empirical behavior of GWT. The compact approximation band captures
substantially more gradient energy than its dimensional fraction, and the
positive permutation energy gaps show that this concentration benefits
from the original local ordering of gradient columns. At the same time,
the complete reconstructed direction remains positively aligned with the
instantaneous gradient in all sampled cases, with the retained detail
coefficients providing a substantial positive contribution. These
properties are consistent across LLaMA-60M and LLaMA-130M and across
GWT-2 and GWT-3, suggesting that the observed structure is stable under
the model-scale and compression-level variations considered here.

\subsection{Study the Effect of $\alpha$}
\begin{figure*}[!th]
    \centering
    \subfloat[\small LLaMA 60M]{\label{fig:ablation_haar1_60_alpha}\includegraphics[width=0.45\linewidth,height=0.25\textwidth]{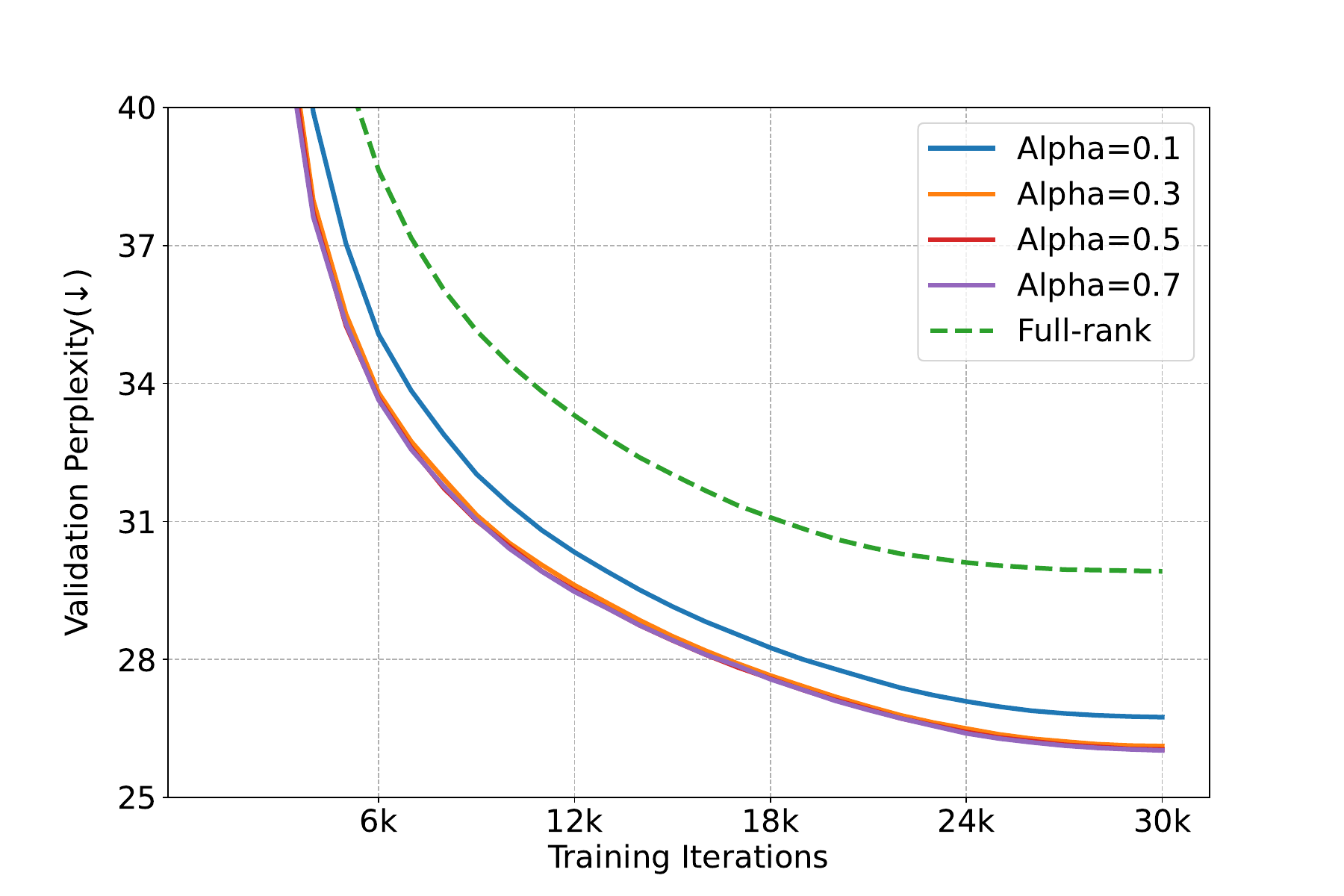}}
    \subfloat[\small LLaMA 130M]{\label{fig:ablation_haar1_130_alpha}\includegraphics[width=0.45\linewidth,height=0.25\textwidth]{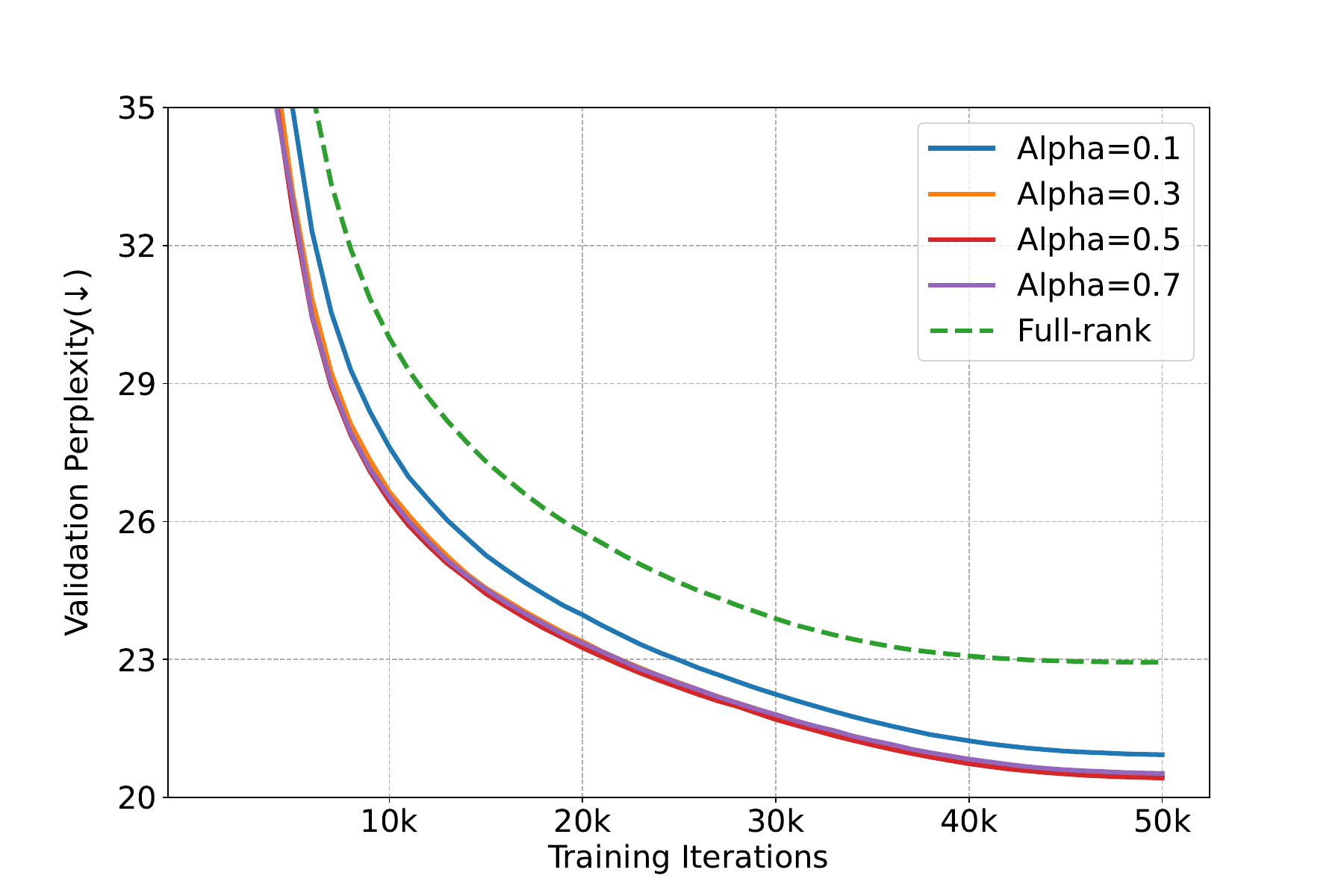}}\\
    \caption{\textbf{Study the effects of $\alpha$ on pre-training LLaMA 60M and 130M models.}} \label{fig:ablation_haar1_60_130m}
\end{figure*}

Similar to other memory-efficient optimizers \cite{zhao2024galore,zhu2024apollosgdlikememoryadamwlevel_apollo,Chen2024FiraCW}, GWT introduces an additional hyperparameter $\alpha$ to control the effective learning rate of the GWT module. In this section, we investigate the impact of this hyperparameter on the performance of GWT. We evaluate Adam with a 2-level Haar GWT across different values of $\alpha$ while fixing the learning rate at $lr = 0.01$. As shown in Figure~\ref{fig:ablation_haar1_60_130m}, our results indicate that GWT performance is largely invariant to the choice of $\alpha$. In particular, for $\alpha > 0.1$, the final performance remains stable with no significant differences. Under our module-wise optimizer strategy, most network modules operate with an effective learning rate greater than $0.001$ ($lr \times \alpha$). In contrast, Adam typically requires learning rates smaller than $0.001$ to prevent loss spikes. This highlights the superior robustness of GWT to higher learning rates. Furthermore, in fine-tuning tasks, tuning $\alpha$ effectively corresponds to adjusting the learning rate $lr$ under our strategy. A more detailed study of the interaction between $lr$ and $\alpha$ in fine-tuning settings is provided in the Appendix.

\subsection{Extension to Other Model Families}

In this section, we extend GWT to models beyond pre-training LLaMA, where we pre-train GPT-2 \cite{Radford2019LanguageMAgpt2}, DeBERTa \cite{he2021debertadecodingenhancedbertdisentangled}, and Qwen2.5-0.5B \cite{yang2024qwen2.5} on the C4 dataset with 2.6B tokens. The final validation loss results are presented in Table~\ref{tab:pre_train_gpt_deberta}. We can see that GWT still maintains superior performance in terms of validation loss across the tested models, demonstrating its effectiveness on other LLM architectures and its strength as a choice for LLM pre-training.

\begin{table}[H]
    \centering
    \setlength{\tabcolsep}{12pt}
    \caption{\textbf{Final validation loss on GPT, DeBERTa, and Qwen.}}
    \label{tab:pre_train_gpt_deberta}
    \begin{tabular}{l|ccc}
    \toprule
       Methods  & GPT & DeBERTa & Qwen \\
       \midrule
       Full-rank Adam  & 3.31 & 2.16 & 2.85\\
       GaLore-1/4 & 3.43 & 2.22 & 2.97 \\
       APOLLO & 3.26 & 2.07 & 2.82\\
       \rowcolor{blue4}\textbf{GWT-2} & \textbf{3.22} & \textbf{2.02} & \textbf{2.70}\\
       \bottomrule
    \end{tabular}
\end{table}

\subsection{Memory and Throughput Estimates}\label{app:memory}
In this section, we provide further clarification on the GPU memory usage results reported in the paper. First, Table~\ref{tab:memory_theory_comp} presents the theoretical memory consumption of various methods for comparison.

Compared to low-rank methods that reduce memory overhead by controlling the rank, our Gradient Wavelet Transform (GWT) method does not require additional storage for projection matrices. In our estimates, GWT requires no persistent projection matrix and can use less optimizer-state memory than the compared low-rank methods at matched compression settings.

In actual training scenarios, GPU memory usage is influenced by multiple factors, including model size, batch size, optimizer states, PyTorch's memory fragmentation management, and whether training is conducted on a single GPU or across multiple GPUs. To focus on core differences, we estimate GPU memory usage for the model weights and Adam optimizer states using bfloat16 precision. For example, the weights of the LLaMA 60M model occupy approximately 0.11 GB of GPU memory, while the Adam optimizer states require about 0.23 GB. Since GWT primarily targets reducing optimizer memory usage, the weight memory footprint under Adam with GWT remains unchanged at 0.12 GB.

As a representative example, the LLaMA-60M model contains approximately 58 million parameters. Using BF16 precision (2 bytes per parameter), this yields a model memory footprint of approximately 0.11 GB. The Adam optimizer maintains two auxiliary states—the first-order moment $M$ and second-order moment $V$—each the same size as the model parameters, resulting in an additional 0.23 GB. The total memory usage with Adam is thus approximately 0.34 GB. For our GWT method, the computation formula depends on the number of parameters that use GWT and those that do not. Specifically, for LLaMA-60M, there are 25.3M parameters that use GWT to save memory, and the remaining 32.77M parameters are updated using Adam. In the case of using GWT with level $l=2$, the optimizer states that the parameters are calculated as follows:
\begin{itemize}
    \item The optimizer state for the GWT parameters:
    $$25.3\ \text{M}\times2\ \text{Bytes}\times2/4=25.3\ \text{MBytes}.$$
    \item  The optimizer state for the Adam parameters:
    $$32.77\ \text{M}\times2\ \text{Bytes}\times2=131.08\ \text{MBytes}.$$
    \item The memory occupied by the model parameters:
    $$116.14\ \text{MBytes}.$$
\end{itemize}
Therefore, the estimated training memory overhead with GWT-2 is
\begin{equation}
    \nonumber
    \begin{aligned}
        &25.3\ \text{MBytes}+131.08\ \text{MBytes}+116.14\ \text{MBytes}\\&=272.52\ \text{MBytes}s\approx0.27\ \text{GBytes}
    \end{aligned}
\end{equation}

Besides the memory estimate, we also report the training throughput as a complement to the training speed report. For example, we pre-train LLaMA 60M with a total batch size of 512 and a max sequence length of 256 in 10,000 iterations. This is equivalent to a total token size of $512\times256=131072\approx131\text{K}$ and a total of $131072 \times 10000=1310720000\approx1.3\text{B}$ tokens during the whole training process. In Tables \ref{tab:throughput_60m_130m_350m} and \ref{tab:pre-llama_hyperparameters}, we provide the training time and the number of GPUs used. Training LLaMA60M with our Haar-2 on four GPUs takes approximately 0.99 hours. Hence, we can compute the token throughput per GPU per second as $1310720000 / (0.99 \times 3600 \times 4)\approx91.9\text{Ktokens/s}$ 

\begin{table*}[!ht]
    \centering
    \setlength{\tabcolsep}{9.5pt}
    \renewcommand{\arraystretch}{1.15}
    \caption{\textbf{Memory estimate of weight parameters/optimizer states.} $^*$ indicates results published in prior works.}
    \label{tab:memory_estimate}
    \begin{tabular}{l|ccccc}
    \toprule
        Methods & 60M & 130M &350M & 1.3B  \\
        \midrule
        Full-Rank &  0.11G/0.23G & 0.26G/0.51G & 0.68G/1.37G & 2.60G/5.20G  \\
        MUON & 0.11G/0.19G & 0.26G/0.38G & 0.68G/0.86G & 2.60G/3.61G\\
        \midrule
        GaLore-1/4 & 0.11G/0.17G & 0.26G/0.32G & 0.68G/0.70G & 2.60G/2.16G \\
        APOLLO-1/4 & 0.11G/0.17G & 0.26G/0.32G & 0.68G/0.70G & 2.60G/2.16G \\
        \rowcolor{blue4}GWT-2 & 0.11G/0.16G  & 0.26G/0.29G & 0.68G/0.56G & 2.60G/1.81G \\
        \midrule
        GaLore-1/8 & 0.11G/0.15G & 0.26G/0.27G & 0.68G/0.55G & 2.60G/1.55G \\
        APOLLO-1/8 & 0.11G/0.15G & 0.26G/0.27G & 0.68G/0.55G & 2.60G/1.55G \\
        \rowcolor{blue4}GWT-3 & 0.11G/0.14G & 0.26G/0.25G & 0.68G/0.41G & 2.60G/1.20G \\
        \midrule
        Low-Rank$^{*}$  & 0.08G/0.17G  & 0.18G/0.37G & 0.36G/0.72G & 1.19G/2.38G\\
        LoRA$^{*}$  & 0.20G/0.17G & 0.44G/0.37G & 1.04G/0.72G & 3.79G/2.38G\\
        ReLoRA$^{*}$  & 0.20G/0.17G & 0.44G/0.37G & 1.04G/0.72G& 3.79G/2.38G\\
         \bottomrule
    \end{tabular}
\end{table*}  

\textbf{GWT level $l$.} We investigate the effects of the GWT level hyperparameter \(l\), with the corresponding validation perplexity curves shown in Figure \ref{fig:learning_curve_level}. We found that the value of $l$ has little impact on the final experimental results, and lower values of $l$ are only slightly better than others. One question that arises from the experimental results is whether the approximation coefficients are not crucial when using wavelets to compress gradients in LLM training, but rather the detail coefficients. The main impact of $l$ seems to be on memory usage and throughput. The throughput results and optimizer memory estimate for different GWT levels are summarized in Table \ref{tab:throughput_60m_130m_350m}, demonstrating the trade-off between memory efficiency and model performance. Higher wavelet levels improve memory usage, but this may come at the cost of slightly higher validation perplexity and slower training speed.
\begin{table}[!th]
    \centering
    \setlength{\tabcolsep}{7pt}
    \renewcommand{\arraystretch}{1.1}
    \caption{Estimated optimizer memory usage and token throughput (per GPU) with different GWT levels. Higher GWT levels lead to slower training token throughput but smaller memory usage.}
    \label{tab:throughput_60m_130m_350m}
    \resizebox{0.6\linewidth}{!}{\begin{tabular}{l|cc|cc|cc}
    \toprule
        \multirow{2}{*}{Methods} & \multicolumn{2}{c}{\textbf{60M}} & \multicolumn{2}{c}{\textbf{130M}} & \multicolumn{2}{c}{\textbf{350M}} \\
       & Memory & Tokens/s  & Memory & Tokens/s  & Memory & Tokens/s \\
    \midrule
        GWT-1 & 0.18G & 95.8K & 0.36G & 46.0K & 0.86G& 12.7K\\
        GWT-2 & 0.16G & 91.9K & 0.29G & 45.1K & 0.56G& 12.1K\\
        GWT-3 & 0.14G & 90.1K & 0.24G & 44.8K & 0.41G& 11.5K\\
        GWT-4 & 0.13G & 84.2K & 0.21G & 43.8K & 0.33G& 10.8K\\
        GWT-5 & 0.13G & 83.5K & 0.20G & 41.5K & 0.30G &9.25K \\
         \bottomrule
    \end{tabular}
    }
\end{table}

\begin{figure*}[!th]
    \centering
    \subfloat[\small LLaMA 60M]{\label{fig:ablation_adam_alpha_60m}\includegraphics[width=0.45\linewidth,height=0.28\textwidth]{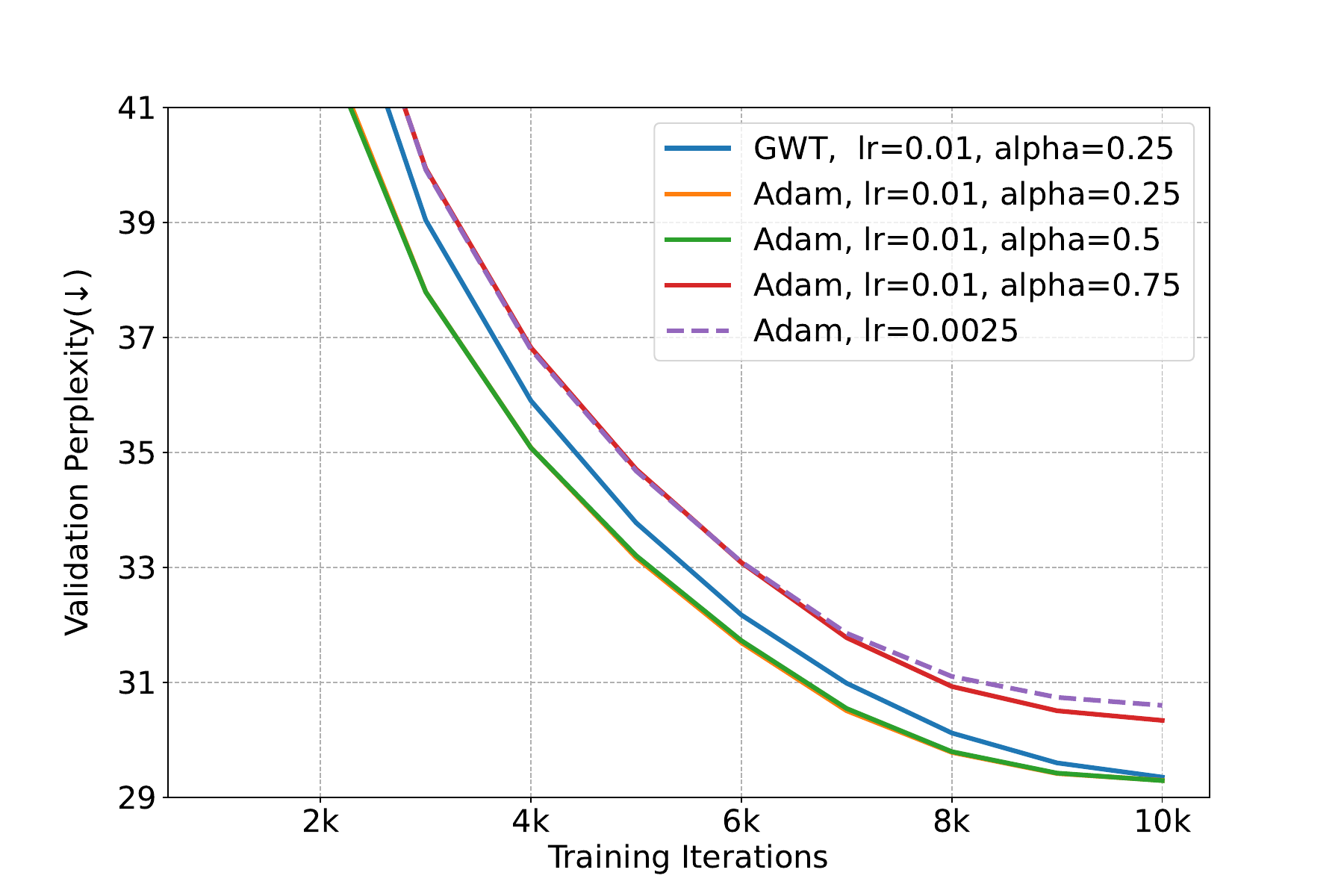}}
    \subfloat[\small LLaMA 130M]{\label{fig:ablation_adam_alpha_130m}\includegraphics[width=0.45\linewidth,height=0.28\textwidth]{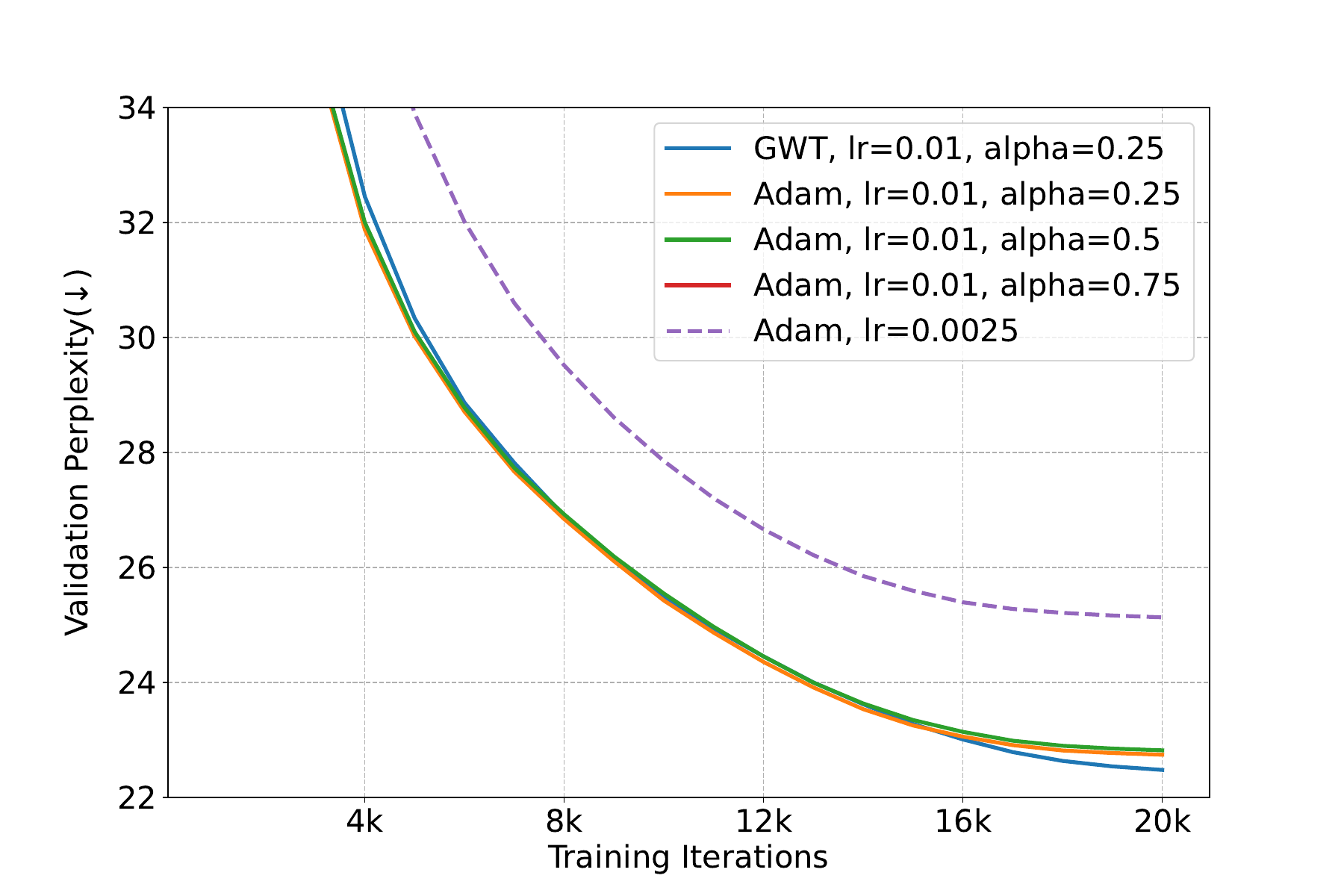}}\\
    \caption{\textbf{Applying module-wise learning rate to Adam optimizer.} We can observe that applying a module-wise split learning rate in Adam leads to better convergence speed and final results compared to using a uniform learning rate across all modules.}\label{fig:ablation_adam_alpha}
\end{figure*}

\subsection{Adam with Module-wise Learning Rate}
One key learning rate strategy employed in GWT is the assignment of different learning rates to distinct parameter modules. Specifically, the overall learning rate is partitioned such that the Attention and MLP modules are updated using a scaled learning rate of $lr \times \alpha$, while the rest of the model uses the base learning rate $lr$. This modular learning rate strategy is commonly adopted in memory-efficient training algorithms, including LoRA \cite{hu2021lora}, GaLore \cite{zhao2024galore}, Fira \cite{Chen2024FiraCW}, and APOLLO \cite{zhu2024apollosgdlikememoryadamwlevel_apollo}. We incorporate this strategy into GWT as a standard configuration. The method of using different learning rates for different network modules in the Adam optimizer has been studied in related works \cite{Everett2024ScalingEAlr, Genet2024CaAdamIAlr}. Here, we provide a supplement as an additional ablation experiment.

In this section, we first apply the same modular learning rate scheme to the Adam optimizer to examine whether it also benefits from this design, and to evaluate whether GWT can still match or surpass Adam’s performance, as demonstrated in the main text. Following standard practice in other memory-efficient methods, we use a learning rate of $lr \times \alpha$ for the Attention and MLP modules, while keeping $lr$ for the remaining modules. To measure the benefit of this configuration, we compare it with Adam using a unified learning rate of 2.5e-3, selected from the range [1e-3, 2.5e-3, 5e-3, 7.5e-3, 1e-2] based on best performance. The learning curves are shown in Figure~\ref{fig:ablation_adam_alpha}.

Experimental results show that, under our setup, Adam significantly benefits from the modular learning rate strategy, achieving clearly better performance than with a single global learning rate. Even with this enhanced setting, GWT achieves a final validation perplexity (PPL) close to that of Full-Adam. This partly explains why many memory-efficient training methods can outperform full-rank Adam—the modular learning rate strategy plays a key role. It also suggests that Attention and MLP modules may require smaller learning rates than other parts of the model. These findings raise an important question: Do different modules in LLMs inherently demand distinct learning rates? And could explicitly integrating modular learning rate strategies into Adam lead to further gains?

\section{Discussion}\label{app:discussion}

GWT and GaLore~\cite{zhao2024galore} both reduce the persistent optimizer states associated with matrix gradients, but they do so in different ways. GaLore updates a low-dimensional projected gradient and discards the component outside the selected subspace. GWT instead applies an orthogonal transform, retains the full coefficient set $(A_t^{(l)},D_t^{(l)},\ldots,D_t^{(1)})$ in the current update, and stores temporal moment states only for $A_t^{(l)}$. The detail coefficients are temporary and therefore do not add persistent optimizer-state memory. Thus, GWT should be viewed as state compression in a fixed transform domain rather than as projected gradient descent.

Here, we outline several intriguing open problems related to GWT:
(a) Wavelets are originally designed for time-series signals and images, raising the question: Can a specialized wavelet transform be developed to handle gradients, which are inherently more disordered and chaotic? Additionally, since the impact of $l$ on the experimental results is minimal, the main effect of higher-order GWT lies in throughput, while offering higher memory efficiency. Therefore, the potential of GWT with higher $l$ values is worth exploring. Given the linear properties of the Haar wavelet in GWT, is it possible to accelerate multi-level wavelet transforms to achieve even higher memory compression ratios, such as 1/64 or 1/128?
(b) The applicability and effectiveness of GWT in other domains, such as vision models \cite{dosovitskiy2020vit} and diffusion models \cite{ho2020denoisingddpm, song2021scorebased}, remain to be validated.

\end{document}